%% file: main.tex
\documentclass{article} 
\usepackage{iclr2027_conference_arxiv,times}

\input{math_commands.tex}

\usepackage{hyperref}
\usepackage{url}

\usepackage{graphicx} 
\usepackage{multirow}
\usepackage{booktabs}
\usepackage{makecell}
\usepackage{caption}
\usepackage{natbib}
\usepackage{placeins}

\title{Controllable Multi-Object Insertion for Detection Data Augmentation}

\author{
Jiyang Huang$^{1*}$ ,
Tianwei Zhang$^{1}$\thanks{These authors contributed equally.}  ,
Xiaoqin Zhang$^{2}$,
Ling Shao$^{3}$,
Shijian Lu$^{1}$ \thanks{Corresponding author.} \\
$^{1}$Nanyang Technological University\\
$^{2}$Zhejiang University of Technology\\
$^{3}$University of the Chinese Academy of Sciences\\
}

\iclrfinalcopy 
\begin{document}

\maketitle

\begin{abstract}
Controllable object insertion provides a targeted approach to augmenting object detection datasets that require category--count fidelity, annotation alignment, and source-content preservation beyond visual realism. 
Several approaches have been proposed to address this grand challenge, but most suffer from different constraints in precise category–count requests and visual consistency, limited coordination across categories and instances, and poor localization of multiple objects of specific category and count.
We design SceneP2F, a two-stage trainable framework that connects scene-aware joint planning with layout-aligned generation leveraging explicit instance-level layouts. SceneP2F consists of two novel designs. The first is \textsc{ScenePlan} which allocates one class-labeled query per requested instance, jointly predicts scene-compatible object centers, and separately estimates box sizes conditioned on the planned centers and scene context. The second is \textsc{SceneAlign-Fill} which combines geometry- and appearance-aware conditioning, box--class-guided regional attention, and an annotation-region-weighted flow-matching objective to render the planned objects while limiting unintended changes to source content. 
Extensive experiments on three challenging benchmarks UAVDT, VisDrone, and BDD100K show that SceneP2F outperforms representative general-purpose editors and object-insertion methods consistently, demonstrating the value of coordinating instance-level planning and generation for category-targeted detection augmentation.

\end{abstract}

\section{Introduction}
Object-detection data often suffer from a long-tail class distribution, motivating controllable object insertion as a targeted approach to adding selected instances to real images.
Rather than collecting and annotating entirely new scenes, object insertion preserve existing backgrounds and annotations while enriching the training distribution with selected instances.
To make such augmentation effective, the inserted objects must be visually consistent with the source scene and occupy plausible locations, while content outside the intended editing regions is preserved.
Reliable control over object categories and counts is equally important for achieving reliable augmentation benefits.

Existing methods have demonstrated the feasibility of object insertion and its utility for downstream tasks.
General-purpose image editors, such as GPT-Image-2.5, FLUX.2~\citep{blackforestlabs2025flux2}, and Qwen-Image-Edit~\citep{wu2025qwenimage}, provide strong generative priors for instruction-guided editing. More directly relevant to our task, task-specific object-insertion methods~\citep{yang2024addsd,zhao2024diffree,tewel2025addit} improve
controllability through specialized training or inference-time guidance. 
However, general-purpose editors may transfer poorly to specialized domains, limiting
their utility for detection augmentation, while existing object-insertion
methods often rely on free-form requests without instance-level constraints for
coordinating new objects with the scene, resulting in unpredictable placement and unreliable category count control. Some methods~\citep{yeh2026xplanner,pmlr-v306-hou26a,petersen2025sceneaware} explicitly predict or sample the spatial layout before image synthesis, improving the plausibility of candidate insertion locations. But they still are not able to jointly achieve precise multi-object control, visual consistency with the source scene, plausible placement, and background preservation, potentially limiting the benefits of downstream data augmentation.

We propose SceneP2F, a two-stage planning--generation framework for \emph{scene-conditioned, category- and count-controlled multi-object insertion} in challenging downstream domains. Given a source scene, the objective of SceneP2F is to add the specified number of \emph{new} instances at plausible locations and scales while maintaining visual consistency and preserving existing content, with the ultimate goal of enabling effective object-level augmentation for downstream detection.

SceneP2F comprises two key designs, namely, \textbf{ScenePlan} and \textbf{SceneAlign-Fill}. \textbf{ScenePlan} decomposes box prediction into size and position components while retaining their conditional dependence, conditioned on the source image, existing annotations, and admissible insertion regions.
It jointly models dependencies among the requested instances and incorporates scene context and geometric feasibility constraints to produce a scene-compatible layout with the requested cardinality.
Given this layout, \textbf{SceneAlign-Fill} jointly renders the planned objects using geometry and appearance conditioning. Regional attention guided by box--class tokens associates each planned region with its target category, while an annotation-region-weighted flow-matching objective emphasizes generation within the designated regions.
These mechanisms encourage the synthesized instances to follow the planned boxes and remain visually compatible with the source scene, while limiting unintended changes outside the editing regions.
Together, the two stages connect explicit category--count constraints at the layout level with instance-level control during image synthesis.

\begin{figure}
    \centering
    \includegraphics[width=1.0\linewidth]{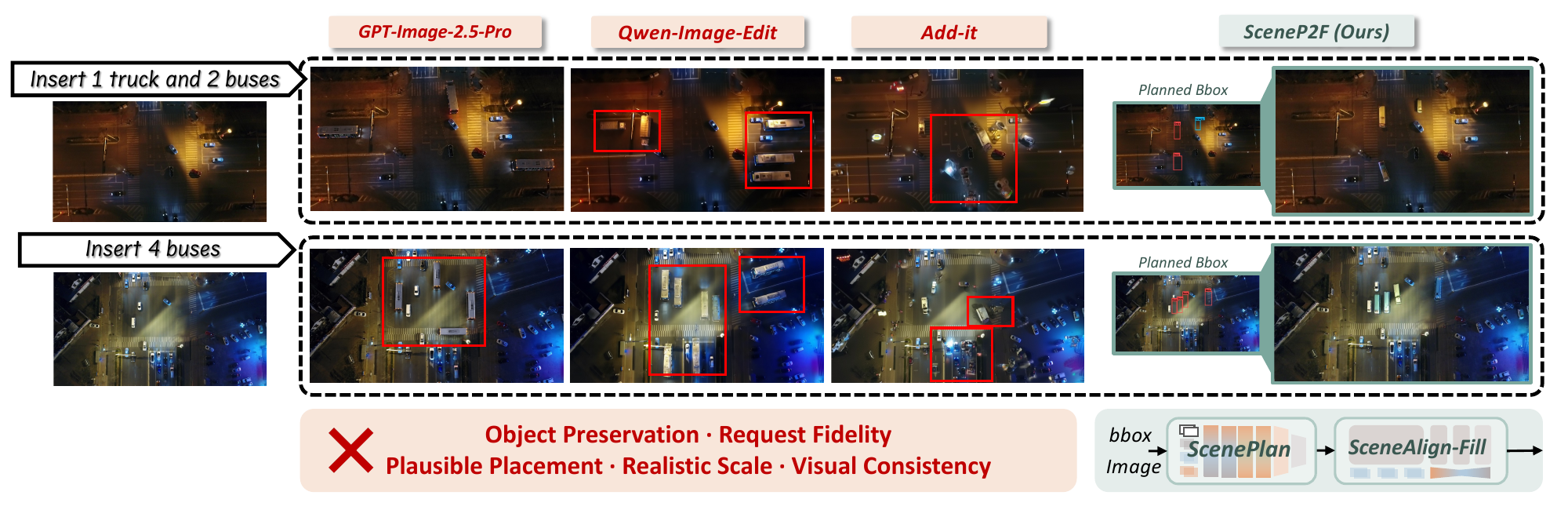}
    \caption{Given a source image and an object insertion request, representative image editing and object insertion approaches may fail to satisfy the requested categories or counts, alter existing objects, or generate implausible placements, scales, and appearances, as highlighted by the red boxes. SceneP2F, a two-stage framework comprising ScenePlan and SceneAlign-Fill, first predicts a scene-compatible, class-labeled box layout and then jointly synthesizes the requested instances within the planned regions, aiming to improve those questions in multi-object insertion task.}
    \label{fig:first_figure}
\end{figure}

We evaluate our framework on two challenging UAV object detection benchmarks, UAVDT~\citep{du2018unmannedaerialvehiclebenchmark} and VisDrone~\citep{zhu2021detection}, with BDD100K~\citep{yu2020bdd100kdiversedrivingdataset} providing a complementary driving-scene benchmark.
Our task-aligned evaluation suite covers two dimensions:
\emph{request fidelity}, which measures adherence to the requested categories and counts of newly inserted instances, and
\emph{scene fidelity}, which assesses insertion naturalness and preservation of the original scene and existing objects.
Comparisons against FLUX.2-dev, Qwen-Image-Edit-2511, Add-SD, and Add-it suggest that SceneP2F improves category–count adherence and provides competitive scene consistency under our evaluation protocol.
SceneP2F achieves the lowest mTCE (Mean Total Count Error) across all three datasets, with particularly large reductions over the best competing baseline on UAVDT (2.744 to 0.817) and BDD100K (5.523 to 1.924). It also improves Outsides PSNR by 5.49, 3.94, and 3.33 dB on UAVDT, VisDrone, and BDD100K, respectively, while achieving the highest DINO similarity and CLIP-I across all three benchmarks.
Adding the resulting synthetic images to detector training improves bus $\mathrm{AP}_{50}$ from 22.9 to 31.0 and truck $\mathrm{AP}_{50}$ from 7.2 to 12.2, demonstrating the potential value of category-targeted insertion for detection augmentation.

Our main contributions are summarized as follows:
(1) We introduce SceneP2F, a trainable framework for annotated detection data generation that integrates category--count control, object-size modeling, appearance alignment, and source-content preservation through explicit instance-level layouts;
(2) We propose \textsc{ScenePlan} to jointly plan scene-compatible layouts through coupled size and position modeling. \textsc{SceneAlign-Fill} uses box-class-guided regional attention and an annotation-region-weighted flow-matching objective to align generated objects with the planned instances while preserving source content;
(3) We show that SceneP2F-generated images and corresponding annotations improve detection of long-tail categories on UAVDT, demonstrating the potential utility of our framework for targeted data augmentation.

\section{Related Work}

\textbf{General-Purpose Image Editing and Object Addition.}
General-purpose Image Editing models, including Qwen-Image-Edit~\citep{wu2025qwenimage}, FLUX.2 [dev]~\citep{blackforestlabs2025flux2}, GLM-Image~\citep{zai2026glmimage}, BAGEL~\citep{deng2025bagel}, and OmniGen~\citep{xiao2025omnigen}, support flexible text- and image-conditioned generation and editing. 
Object-addition methods can be organized by their dominant control interface.
\textit{Mask-free editing} leaves placement to the model: InstructPix2Pix~\citep{Brooks_2023_CVPR} establishes instruction-guided editing; Diffree~\citep{zhao2024diffree} and Paint by Inpaint~\citep{Wasserman_2025_CVPR} learn insertion from removal-based supervision; and Add-it~\citep{tewel2025addit} uses training-free attention control.
\textit{Spatially grounded editing} uses user-specified regions in SmartBrush~\citep{xie2023smartbrush} and ObjectAdd~\citep{zhang2025objectaddaddingobjectsimage}, or instruction-derived plans in X-Planner~\citep{yeh2026xplanner} and MIRAGE~\citep{liu2026mirage}, with the latter supporting multi-instance editing.
\textit{Reference-conditioned composition} transfers exemplar content: Paint by Example~\citep{yang2023paintbyexample} and ObjectStitch~\citep{song2023objectstitch} perform diffusion compositing; AnyDoor~\citep{chen2024anydoor} emphasizes identity and detail preservation; and Insert Anything~\citep{song2026insertanything}, FreeInsert~\citep{zhang2025freeinsert}, and Multitwine~\citep{tarres2025multitwine} extend prompting flexibility, geometric and style control, and joint multi-object composition, respectively.

\textbf{Detection-Oriented Synthetic Data Generation.}
\textit{Compositional augmentation} constructs training samples by transferring foreground instances, as in Copy-Paste~\citep{Dwibedi_2017_ICCV, Dvornik_2018_ECCV} and Simple Copy-Paste~\citep{Ghiasi_2021_CVPR}. X-Paste~\citep{pmlr-v202-zhao23f} and \textit{Stable Diffusion for Aerial Object Detection}~\citep{jian2023stablediffusionaerialobject} extend this strategy with diffusion-generated foregrounds.
\textit{Layout-conditioned synthesis} generates images from categories and spatial constraints, as in LayoutDiffusion~\citep{Zheng_2023_CVPR} and GLIGEN~\citep{Li_2023_CVPR}. GeoDiffusion~\citep{chen2023geodiffusion}, ODGEN~\citep{zhu2024odgen}, AeroGen~\citep{tang2024aerogen}, and UAVGen~\citep{li2026uavgen} extend this paradigm to detection-oriented data generation, while PlanGen~\citep{He_2025_ICCV} incorporates global layout planning. DatasetDM~\citep{wu2023datasetdm} instead generates images and predicts their perception annotations.
\textit{Task-aware synthesis} improves training utility through data-distribution modeling or detector feedback. Add-SD~\citep{yang2024addsd} uses frequency-aware category sampling; AeroGen~\citep{tang2024aerogen} models and filters remote-sensing layouts; and UAVGen~\citep{li2026uavgen} combines class prototypes and focal regions with detector-based annotation refinement. DetDiffusion~\citep{wang2024detdiffusion} uses detector-derived difficulty attributes, whereas \textit{Unbiased Object Detection Beyond Frequency}~\citep{cai2026unbiased} adapts generation to representation gaps using detector feedback. These strategies address synthetic-data composition and utility, complementing our focus on per-image category--count control.

\textbf{Scene-Conditioned Object Insertion and Layout Planning.}
Scene-conditioned insertion ranges from implicit placement to explicit geometric planning. Add-SD~\citep{yang2024addsd} learns insertion from object-removal pairs; \textit{Imagining the Unseen}~\citep{yun2024imagining} predicts class-conditioned 2D boxes for inpainting; Scene-Aware Location Modeling~\citep{petersen2025sceneaware} samples placements using scene geometry and dataset statistics; MonoPlace3D~\citep{Parihar_2025_CVPR} learns distributions over plausible 3D boxes; and PDA~\citep{pmlr-v306-hou26a} combines physics-grounded planning with appearance adaptation.
Joint layout modeling also appears in traffic simulation: SceneGen places actors autoregressively, while Scenario Diffusion and SceneControl use diffusion to generate actor configurations under scene constraints~\citep{Tan_2021_CVPR,pronovost2023scenario,scenecontrol2024}. Scenario Dreamer jointly generates lane graphs and actor boxes, and ScenarioControl supports text- or image-conditioned vectorized scenes~\citep{Rowe_2025_CVPR,scenariocontrol2026}.
For detection-oriented insertion, the challenge extends beyond plausible placement to maintaining instance-level consistency between a complete category--count request and the generated image. Our framework uses an explicit, class-labeled 2D layout to connect scene-aware joint planning with layout-aligned generation, promoting request fidelity, image-annotation alignment, and source-content preservation.

\section{Methodology}
\label{sec: methods}

\subsection{Framework Overview}

Given an image $I$, its existing object configuration $\mathcal{C}$, and a structured category--count request $\mathcal{R}=\{(c_k,n_k)\}_{k=1}^{K}$, our goal is to generate an edited image $\hat{I}$ and insertion annotations $b^*$ containing exactly $n_k$ new instances of each category $c_k$. 

\begin{figure}[!htbp]
    \centering
    \includegraphics[width=\linewidth]{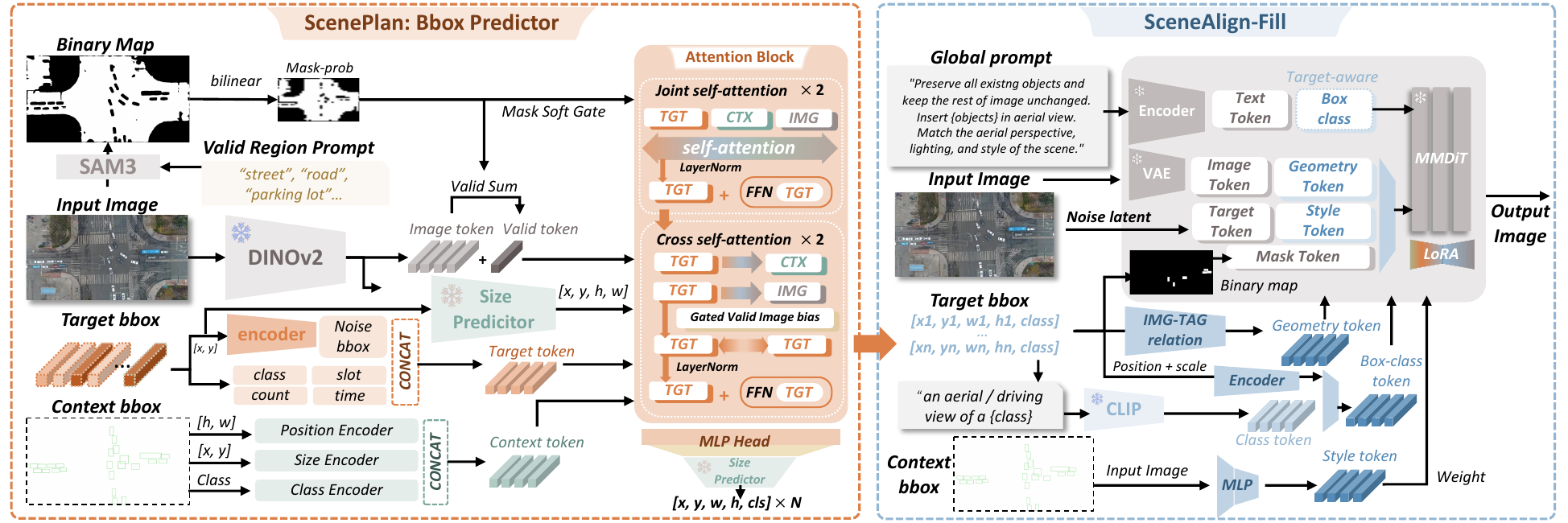}
    \caption{Overview of the SceneP2F framework, comprising ScenePlan (left) and SceneAlign-Fill
    (right). ScenePlan first jointly predicts object centers from image features, existing boxes, and category--count queries, while a size predictor estimates the corresponding box extents. Given the resulting class-labeled layout, SceneAlign-Fill uses geometric, categorical, and scene-style conditioning to synthesize the requested objects while limiting changes outside the target regions.}
    \label{fig:framework_overview}
\end{figure}

We address these requirements in a two-stage \textbf{SceneP2F} method. As shown in Fig.~\ref{fig:framework_overview}. the framework consists of two modules: ScenePlan and SceneAlign-Fill. Firstly, \textbf{ScenePlan} expands $\mathcal{R}$ into category-specific queries, jointly denoises their centers conditioned on $I$, $\mathcal{C}$, and the validity mask, and employs a scene-aware size predictor to realize them as complete boxes. Finally, \textbf{SceneAlign-Fill} inserts requested objects by rendering the planned objects with scene-aware conditioning and box-restricted regional attention, ensuring appearance consistency, and source-scene preservation. Details of SceneP2F training among three datasets are provided in Appendix~\ref{sec:appendix_SceneP2F_training_and_inference}.

\subsection{ScenePlan: Planned Scene-Conditioned Box Planning}

We formulate insertion-layout planning as category and count constrained set prediction: given a background image $I$, existing context annotations $\mathcal{C}=\
\{(\mathbf{b}^{\mathrm{ctx}}_j,c^{\mathrm{ctx}}_j)\}_{j=1}^{N_{\mathrm{ctx}}}$, and a request $\mathcal{R}=\{(c_k,n_k)\}_{k=1}^{K}$, our ScenePlan predicts a set of target boxes:
\begin{equation}
    \widehat{\mathcal{L}}
    =
    \left\{
        \left(\widehat{\mathbf{b}}^*_i,
        c^{\mathrm{tgt}}_i\right)
    \right\}_{i=1}^{N_{\mathrm{tgt}}}, \quad
    \sum_{i=1}^{N_{\mathrm{tgt}}}
    \left[
        c^{\mathrm{tgt}}_i=c_k
    \right]
    =
    n_k,\quad
    k=1,\ldots,K.
\end{equation}

Each box is parameterized as $\widehat{\mathbf{b}}^*_i=(\hat{x}_i,\hat{y}_i,\hat{w}_i,\hat{h}_i)$, where $(\hat{x}_i,\hat{y}_i)\in[0,1]^2$ and $(\hat{w}_i,\hat{h}_i)\in(0,1]^2$ denote its normalized center and extent, respectively. We expand the category--count request into \(N_{\mathrm{tgt}}=\sum_{k=1}^{K}n_k\) active queries with fixed category assignments, such that the category multiplicities and output cardinality are determined by construction at the layout-representation level.

\begin{figure}
    \centering
    \includegraphics[width=1.0\linewidth]{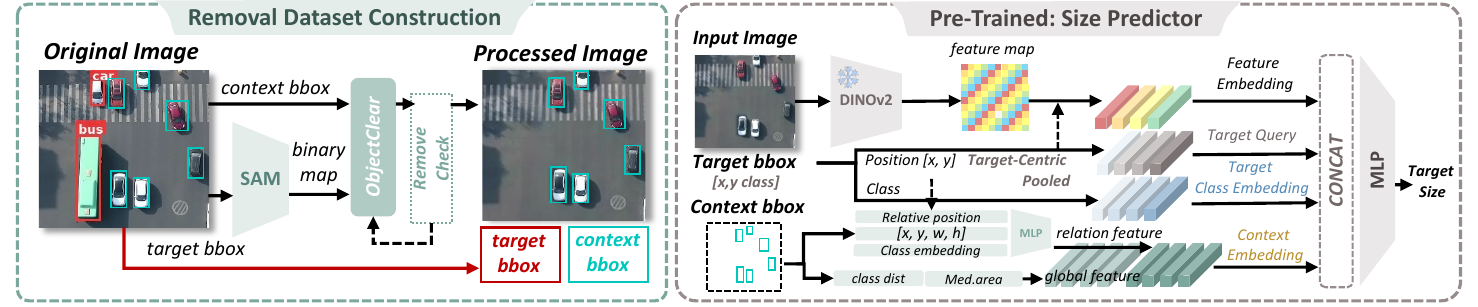}
    \caption{Removal dataset construction (left) and pretrained size predictor (right). The removal-data construction pipeline uses SAM2 to generate masks and ObjectClear to remove the target objects, while a verification module triggers iterative reprocessing when the check fails. Size predictor sub-model estimates target-box dimensions from image features, context boxes, and input center.}
    \label{fig:removal-size}
\end{figure}

\textbf{Removal Dataset Construction.}
ScenePlan is trained on our task-specific dataset. Inspired by works \citep{canberk2024erasedraw,Wasserman_2025_CVPR,winter2024objectdrop,li2025rorem,winter2025objectmate,yu2025objectmover} on object insertion and spatial prediction methods trained by their removal-pair datasets, we construct ours by mapping entire annotations to a unified label space and splitting valid instances into target to remove and context boxes to retain, and further sampling and filtering details are provided in Sec.~\ref{preparation details}. As shown in Fig.~\ref{fig:removal-size}, we generate slightly dilated target masks with SAM2~\citep{ravi2024sam2}, protect annotated non-target regions, and apply ObjectClear~\citep{zhao2026objectclear}, optionally with a second refinement pass to filter unsuccessful edits by using edge-density and image-difference criteria. 

\textbf{Decoupled size-coords prediction.} Before training the main ScenePlan model, we first design and pre-train a sub-model $G_{\phi}$ to estimate the width and height of each target box, which is illustrated in Fig.~\ref{fig:removal-size}. It takes the image features, context-box tokens, and the planned center query as input and uses lightweight feature-fusion layers followed by an MLP regressor to predict
\begin{equation}
    \widehat{\mathbf{s}}_i
    =
    G_{\phi}
    \left(
        I,\mathcal{C},c_i,
        \widehat{\mathbf{z}}_{0,i}
    \right),
    \qquad
    \widehat{\mathbf{b}}_i
    =
    \left(
        \widehat{\mathbf{z}}_{0,i},
        \widehat{\mathbf{s}}_i
    \right),
    \label{eq:size_prediction}
\end{equation}
where $c_i$ is the request class, $\widehat{\mathbf{z}}_{0,i}=(\hat{x}_i,\hat{y}_i)$ and $\widehat{\mathbf{s}}_i=(\hat{w}_i,\hat{h}_i)$. We initially attempted to predict complete bounding boxes jointly, but found the predictions unstable, particularly for width and height, which often deviated substantially from the ground-truth extents. This observation motivated us to decouple center prediction from size estimation. We provide ablation of the model in Appendix.\ref{sec:appendix-size-predictor}.

\textbf{Plausible Region Control.} ScenePlan incorporates scene-conditioned positional validity through spatial-validity mask $M_{valid} \in [0,1] $ generated by SAM3 ~\citep{carion2025sam3segmentconcepts}. Dataset-specific prompts cover plausible support regions while excluding overly generic terms that may merge semantically distinct surfaces, and details are described in Sec.\ref{Valid region mask construction}. We incorporate the mask in three complementary ways: (1) construct a valid summary token by mask-weighted pooling of image features followed by an MLP: (2) transform the mask into a logarithmic soft bias on target-to-image attention logits; (3) include a region-validity penalty loss illustrated in the next part.

\textbf{ScenePlan Diffusion Model}
ScenePlan encodes $I$ input image using DINOv2 and then projected to imgae token $\tau_{u_k}$. The context object token $\tau_{c_j}$ and target token $\tau_{z_i^0}$ are encoded as
\[
\tau_{c_j} =
\operatorname{MLP}_{xy}(b^{ctx}_{j,xy})
+
\operatorname{MLP}_{size}(b^{ctx}_{j,\log wh})
+
e_{\mathrm{class}}(c^{ctx}_j),
\]
\[
\tau_{z_i^0} =
\operatorname{MLP}_{xy}(z^{tgt}_{t,i})
+ e_t(t)
+ e_{\mathrm{slot}}(i)
+ e_{\mathrm{count}}(N_{tgt})
+ e_{\mathrm{class}}(c^{tgt}_i)
\]
, where \(\mathbf z_{t,i}\) is the noisy center of target query \(i\) at timestep \(t\), and \(e(\cdot)\) denotes a learnable embedding. 
The tokens are processed by a four-layer relation-aware transformer. The first two layers perform joint self-attention over the target, context, and image tokens, whereas the remaining two refine the target tokens through target to-target self-attention and cross-attention to the context and image tokens. The frozen size predictor $G_{\phi}$ provides provisional geometry for relative cross-attention biases and subsequently predicts the final box extents from the refined centers and scene context, decoupling center planning from size estimation while preserving their conditional dependence.

ScenePlan applies diffusion only to the centers of the \(N_{\mathrm{tgt}}\) active target queries. Given their noisy center set \(\mathbf Z_t=[\mathbf z_{t,i}]_{i=1}^{N_{\mathrm{tgt}}}\) at timestep \(t\) and the conditioning information \(\Omega=(I,\mathcal C,\mathcal R,\mathbf{M_{valid}})\), the denoiser jointly predicts the corresponding velocity set using a \(T=100\)-step DDPM:
\begin{equation}
\widehat{\mathbf V}_t=f_\theta(\mathbf Z_t,t,\Omega),\qquad
\mathbf v_{t,i}
=
\sqrt{\bar{\alpha}_t}\boldsymbol{\epsilon}_i
-
\sqrt{1-\bar{\alpha}_t}\mathbf z_{0,i},
\quad
\boldsymbol{\epsilon}_i\sim\mathcal N(\mathbf0,\mathbf I_2),
\label{eq:center_velocity}
\end{equation}
where \(\bar{\alpha}_t=\prod_{s=1}^{t}(1-\beta_s)\). Starting from Gaussian center noise, the standard DDPM reverse process progressively refines the joint center state and outputs the planned center set \(\widehat{\mathbf Z}_0\), obtaining the corresponding size by $G_{\phi}$.

For training objective, we supervise ScenePlan using a velocity-prediction loss, a timestep-weighted set-level coordinate loss, and a validity loss that encourages the recovered centers to lie within the shared admissible region:
\begin{equation}
\mathcal L_{\mathrm{coord}}
=
\mathbb E_{t,\epsilon}
\left[
w_t\,
\mathcal D_{\mathrm{set}}
\left(
\widehat{\mathcal Z}_{0\mid t},
\mathcal Z_0
\right)
\right],
\quad
\mathcal L_{\mathrm{mask}}
=
\mathbb E_t
\left[
\frac{1}{N_{\mathrm{tgt}}}
\sum_i
\left(
1-
\rho
\left(
\mathbf M_{valid},
\widehat{\mathbf b}_{0\mid t,i}
\right)
\right)
\right].
\label{eq:sceneplan_losses}
\end{equation} 
Here, \(w_t\) is an SNR-based timestep weight, 
$\mathcal D_{\mathrm{set}}$ denotes the average Smooth-L1 coordinate loss after class-constrained Hungarian matching between the recovered and ground-truth centers. $\widehat{\mathcal Z}_{0\mid t}
=
\{\widehat{\mathbf z}_{0\mid t,i}\}_{i=1}^{N_{\mathrm{tgt}}}$
denotes the recovered clean-center set and $\rho(\mathbf{M_{valid}},\widehat{\mathbf b}_{0\mid t,i})$ denotes the differentiable fraction of the predicted box covered by the valid-region mask.
The final objective $L_{\mathrm{ScenePlan}}$ is
\begin{equation}
\mathcal L_{\mathrm{ScenePlan}}
=
\mathbb E_{t,\boldsymbol\epsilon}
\left[
\frac{1}{N_{\mathrm{tgt}}}
\sum_i
\left\|
\widehat{\mathbf v}_{t,i}-\mathbf v_{t,i}
\right\|_2^2
\right]
+
\lambda_{\mathrm{coord}}\mathcal L_{\mathrm{coord}}
+
\lambda_{\mathrm{mask}}\mathcal L_{\mathrm{mask}}.
\label{eq:sceneplan_objective}
\end{equation}

In the inference, we apply a deterministic post-process to locally adjust predictions that violate basic boundary or overlap constraints to obtain the final $\hat{b}^*_i$. The requested categories and instance count remain unchanged. Implementations details for each dataset are shown in Appendix~\ref{sec:appendix-post-process}.

\subsection{SceneAlign-Fill: Annotation-Aligned Scene Refinement}
\label{sec:refine_flux_fill}

Given an input scene $I$, a target layout
$\hat{\mathcal L}=\{(b_i^*,c_i)\}_{i=1}^{N_{\mathrm{tgt}}}$, and a text prompt $T$, SceneAlign-Fill synthesizes the requested objects while enforcing category--box alignment, preserving unedited content, and maintaining scene consistency. Since FLUX.1-Fill relies mainly on a shared inpainting mask and a global prompt, it provides limited instance-level control over multi-category,
multi-instance insertion. We address this limitation through scene and annotation image conditioning, box-token regional attention, and an annotation-weighted flow-matching objective.
 
\textbf{Scene-conditioned image conditioning.}
Let $z_{\mathrm{cond}}$ denote the VAE latent representation of $x_{\mathrm{safe}}$. For each image token $p$, we construct the image-side condition
\begin{equation}
    C_p=
    \left[
        z_{\mathrm{cond},p};
        \tau_{m_p};
        \tau_g;
        \tau_s
    \right],
    \label{eq:compact_condition}
\end{equation}
where $\tau_{m_p}$ is the packed inpainting-mask feature, $\tau_g$ is a geometry token, and $\tau_s$ is a scene-level style token. The geometry token encodes the relative position of image token with respect to its associated target box with the its scale. The style token is obtained by aggregating color and latent texture features from visible context objects and is shared across all image tokens. Together, these conditions enable spatially controlled multi-instance synthesis while encouraging the inserted objects to follow the visual consistency. Details of token process are shown in Appendix~\ref{sec:appendix_SceneAlign-Fill}.

\textbf{Box-token regional attention.}
To provide region-specific semantic control, we represent each target annotation with a box token:
\begin{equation}
    u_i =
    f_{\mathrm{box}}
    \left(
        [e_{c_i};\phi(b_i)]
    \right).
    \label{eq:compact_box_token}
\end{equation}
Here, $e_{c_i}$ is obtained by encoding a local prompt for category $c_i$ with the frozen CLIP text encoder, while $\phi(b_i)$ represents the normalized position and scale of box $b_i$. The resulting box tokens are appended to the FLUX text sequence and used for regional attention in the late Transformer blocks: image tokens within $b_i$ attend only to the corresponding box token $u_i$, whereas global text tokens remain accessible to all regions. This
restricts category-specific information to the intended boxes, reducing cross-instance semantic leakage and improving image--annotation alignment.

\section{Experiments}
\subsection{Multi-Object Insertion}
\begin{figure}
    \centering
    \includegraphics[width=1.0\linewidth]{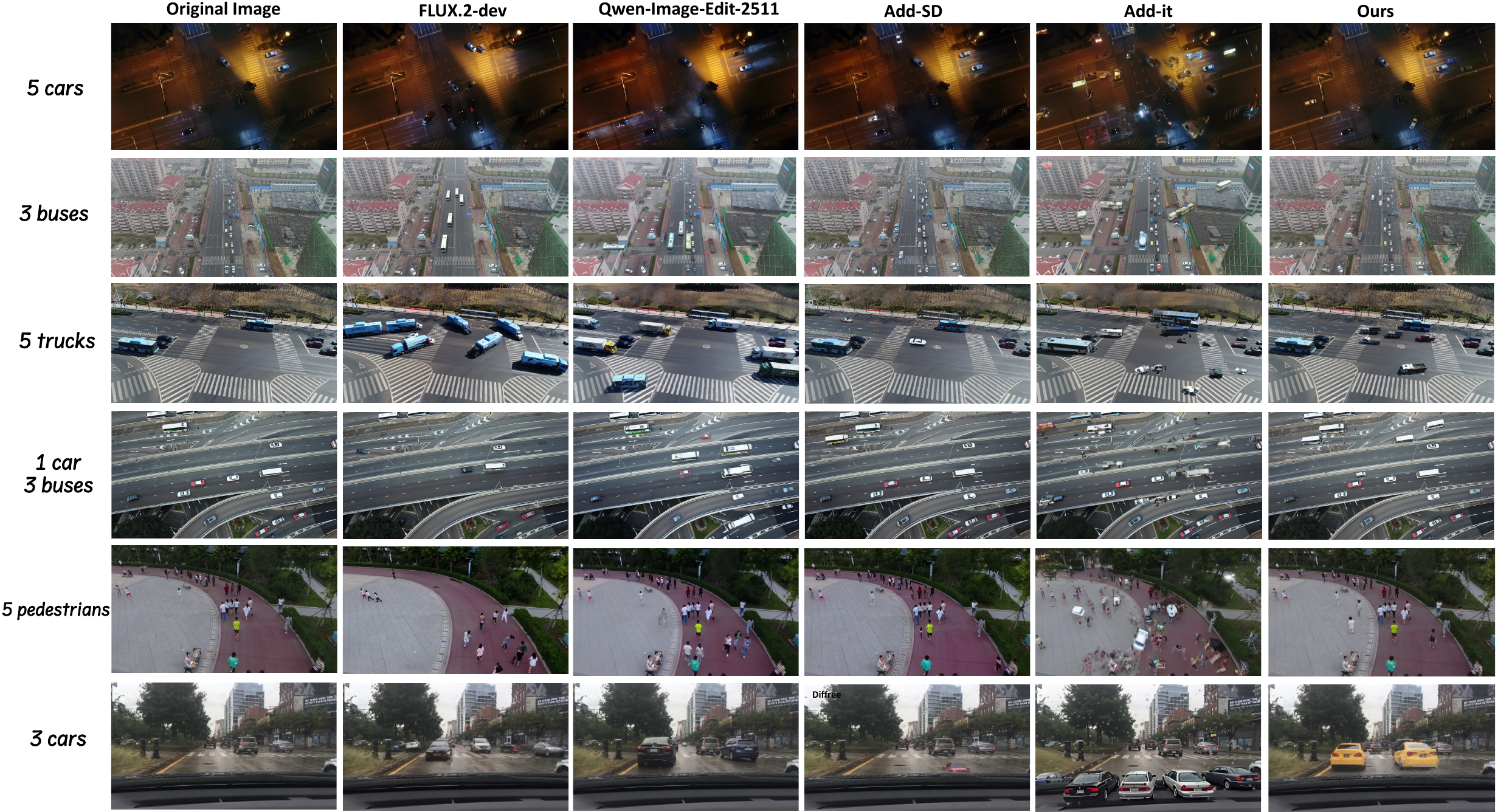}
    \caption{Qualitative comparison of SceneP2F with image-editing models (FLUX.2-dev and Qwen-Image-Edit) and object-insertion baselines (Add-SD and Add-it) on the joint multi-object insertion task. The first three rows present results on UAVDT, the subsequent two rows on VisDrone, and the final row on BDD100K. The leftmost column shows the requested categories and counts. For BDD100K (last row), Diffree replaces Add-SD, which was not trained on this dataset.}
    \label{fig:Object_Insertion_comparison}
\end{figure}

\textbf{Settings.}
We train dataset-specific models on
UAVDT~\citep{du2018unmannedaerialvehiclebenchmark},
VisDrone~\citep{zhu2021detection}, and
BDD100K~\citep{yu2020bdd100kdiversedrivingdataset}. We use the official training
splits for VisDrone and BDD100K and a random 40/10 train/validation fixed sequence
split for UAVDT. All methods are evaluated with the same predefined request
schedules on all 1,609 VisDrone test images and 1,000 sampled images each from
BDD100K and UAVDT. We use category-balanced single- and multi-class requests with $N\in\{1,\ldots,5\}$ and an equal budget for each $N$. All methods use input prompt \textit{``Preserve all existing objects and keep the rest of the image unchanged. Insert \{objects\} into the \{scene type\} driving scene while matching its perspective, lighting, and style,''} where \textit{scene type} is
set to \textit{aerial} or \textit{street-view} according to the dataset.
Further details are provided in Appendix~\ref{sec:appendix-eval-details}.

\textbf{Baseline Methods.}
We compare our method with two groups of baselines. General-purpose editors include FLUX.2 [dev]~\citep{blackforestlabs2025flux2}, evaluated using only its official checkpoint, and Qwen-Image-Edit-2511~\citep{wu2025qwenimage}, evaluated both with its official checkpoint and after fine-tuning on each aerial dataset. Object-insertion baselines include Add-it~\citep{tewel2025addit}, a training-free method built on FLUX.1-dev, and Add-SD~\citep{yang2024addsd}, which is closely related to our task but was originally developed for a different image domain. For each aerial benchmark, we fine-tune Add-it~\citep{tewel2025addit} from pretrained FLUX.1-dev weights and adapt Add-SD~\citep{yang2024addsd} using the corresponding training set. We do not fine-tune the baseline methods on BDD100K~\citep{yu2020bdd100kdiversedrivingdataset}, as its street-view imagery exhibits relatively little domain shift. Appendix.~\ref{sec:appendix-baseline model finetune} illustrates the fine-tuned details. To ensure a fair comparison under the same supervision setting, we also evaluate our model without access to ground-truth bounding-box annotations.

\textbf{Metrics.}
We use dataset-specific GroundingDINO detectors~\citep{liu2023grounding} to identify inserted objects, and we present details of the detector in Appendix~\ref{sec:appendix-detector-gd}. We report three groups of metrics:
(1) \textbf{Request Fidelity}, measured by Mean Total Count Error
(mTCE)$\downarrow$ and Mean Class Count Error (mCCE)$\downarrow$, which quantify
total and class-wise deviations from the requested counts;
(2) \textbf{Visual Consistency}, measured by
LPIPS$\downarrow$, DINO$_{\mathrm{Sim}}$$\uparrow$, and CLIP-I$\uparrow$ for
perceptual distance, semantic similarity, and visual-semantic similarity,
respectively; and
(3) \textbf{Outside Preservation}, measured by PSNR$\uparrow$ and
SSIM$\uparrow$ over unedited regions. Detected and ground-truth boxes define the generated and reference crops, respectively. LPIPS is averaged over crop pairs and images, whereas DINO$_{\mathrm{Sim}}$ and CLIP-I are computed by comparing each generated-crop feature with the image-level mean reference feature and averaging the cosine similarities. More details are presented in Appendix.\ref{sec:appendix-metrics}.

\textbf{Results.} We present visualization comparison in Fig.~\ref{fig:Object_Insertion_comparison}. Table.~\ref{tab:object-insertion-comparison} shows that our method achieves the strongest overall performance. It obtains the lowest mTCE and mCCE among all baselines on the three datasets, with clear gains on UAVDT~\citep{du2018unmannedaerialvehiclebenchmark} ($0.817/1.189$ vs.\ $2.744/2.105$) and BDD100K ($1.924/1.648$ vs.\ $5.523/2.576$), demonstrating more accurate category--count control. Our method also achieves the highest CLIP-I scores on all datasets and consistently lower LPIPS values than the baselines, indicating that the inserted objects are natural and visually consistent with the original scenes. The results show that our method demonstrates good overall performance regardless of whether GT annotations are provided.
We also evaluate the accuracy of insertion count shown in Fig\ref{fig:count_accuracy_graph}, which is calculated the rate of successfully inserting request objects into the image.

\begin{table*}[!t]
\caption{Quantitative comparison of general image-editing and
object-insertion methods on UAVDT, VisDrone, and BDD100K.
The best results are shown in bold, and the second-best are underlined.}
\label{tab:object-insertion-comparison}
\centering
\small

\renewcommand{\arraystretch}{0.88}
\setlength{\tabcolsep}{3.5pt}

\resizebox{\textwidth}{!}{
\begin{tabular}{@{}lllcccccccc@{}}
\toprule

\multirow{2}{*}{\bfseries Dataset}
&
\multirow{2}{*}{\makecell{\bfseries Method\\\bfseries Category}}
&
\multirow{2}{*}{\bfseries Method}
&
\multirow{2}{*}{\makecell{\bfseries Fine-\\\bfseries tuned}}
&
\multicolumn{2}{c}{\bfseries Request Fidelity}
&
\multicolumn{3}{c}{\bfseries Visual Consistency}
&
\multicolumn{2}{c}{\bfseries Outside Preservation}
\\

\cmidrule(lr){5-6}
\cmidrule(lr){7-9}
\cmidrule(lr){10-11}

& & & &
\bfseries mTCE $\downarrow$
&
\bfseries mCCE $\downarrow$
&
\bfseries LPIPS $\downarrow$
&
\bfseries DINO$_{\mathrm{Sim}}$ $\uparrow$
&
\bfseries CLIP-I $\uparrow$
&
\bfseries PSNR $\uparrow$
&
\bfseries SSIM $\uparrow$
\\

\midrule

\multirow{7}{*}{
    \makecell{UAVDT\\
    ~\citep{du2018unmannedaerialvehiclebenchmark}}
}
&
\multirow{3}{*}{\makecell[l]{Image\\Editing}}
&
FLUX.2-dev~\citep{blackforestlabs2025flux2}
&
{}
& 7.001 & 3.076 & 0.5561 & 0.4753 & 0.8131 & 18.9585 & 0.8405
\\

&
&
\multirow{2}{*}{Qwen-Image-Edit~\citep{wu2025qwenimage}}
&
{}
& 5.596 & 2.458 & 0.5588 & 0.4635 & 0.8188 & 20.6050 & 0.8973
\\

&
&
&
$\checkmark$
& 14.959 & 7.603 & 0.5051 & 0.4499 & 0.8452 & 19.3728 & 0.8408
\\

\cmidrule(lr){2-11}

&
\multirow{2}{*}{\makecell[l]{Object\\Insertion}}
&
Add-it~\citep{tewel2025addit}
&
$\checkmark$
& 8.529 & 4.594 & 0.5164 & 0.4508 & 0.8410 & 21.5469 & 0.8987
\\

&
&
Add-SD~\citep{yang2024addsd}
&
$\checkmark$
& 2.744 & 2.105 & 0.5751 & 0.4295 & 0.8356 & 22.2962 & 0.9010
\\

\cmidrule(lr){2-11}

&
\multicolumn{2}{l}{Ours (w/o context bbox)}
&
{}
& \underline{0.916}
& \underline{1.243}
& \textbf{0.4727}
& \underline{0.4901}
& \underline{0.8532}
& \underline{27.7671}
& \textbf{0.9635}
\\

&
\multicolumn{2}{l}{\textbf{Ours}}
&
{}
& \textbf{0.817}
& \textbf{1.189}
& \underline{0.4822}
& \textbf{0.4941}
& \textbf{0.8566}
& \textbf{27.7819}
& \underline{0.9634}
\\

\midrule

\multirow{7}{*}{
    \makecell{VisDrone\\
    ~\citep{zhu2021detection}}
}
&
\multirow{3}{*}{\makecell[l]{Image\\Editing}}
&
FLUX.2-dev~\citep{blackforestlabs2025flux2}
&
{}
& 25.557 & 10.661 & 0.5564 & 0.5098 & 0.8582 & 22.3622 & 0.9181
\\

&
&
\multirow{2}{*}{Qwen-Image-Edit~\citep{wu2025qwenimage}}
&
{}
& 19.864
& 9.841
& 0.5558
& \underline{0.5370}
& 0.8671
& 20.8987
& 0.8861
\\

&
&
&
$\checkmark$
& 39.184 & 19.323 & 0.5602 & 0.5282 & 0.8595 & 18.5221 & 0.7800
\\

\cmidrule(lr){2-11}

&
\multirow{2}{*}{\makecell[l]{Object\\Insertion}}
&
Add-it~\citep{tewel2025addit}
&
$\checkmark$
& 44.760 & 20.926 & 0.5629 & 0.5305 & 0.8623 & 19.8011 & 0.8273
\\

&
&
Add-SD~\citep{yang2024addsd}
&
$\checkmark$
& \underline{7.776}
& \underline{5.249}
& 0.5656
& 0.5304
& 0.8655
& 22.6756
& 0.9148
\\

\cmidrule(lr){2-11}

&
\multicolumn{2}{l}{Ours (w/o context bbox)}
&
{}
& 11.605
& 6.187
& \underline{0.5557}
& 0.5364
& \underline{0.8682}
& \underline{23.5125}
& \underline{0.9221}
\\

&
\multicolumn{2}{l}{\textbf{Ours}}
&
{}
& \textbf{7.510}
& \textbf{5.018}
& \textbf{0.5355}
& \textbf{0.5611}
& \textbf{0.8784}
& \textbf{26.6133}
& \textbf{0.9669}
\\

\midrule

\multirow{5}{*}{
    \makecell{BDD100K\\
    ~\citep{yu2020bdd100kdiversedrivingdataset}}
}
&
\multirow{2}{*}{\makecell[l]{Image\\Editing}}
&
FLUX.2-dev~\citep{blackforestlabs2025flux2}
&
{}
& 7.773
& 3.420
& 0.6178
& 0.4711
& 0.8215
& \underline{27.8150}
& \underline{0.9428}
\\

&
&
Qwen-Image-Edit~\citep{wu2025qwenimage}
&
{}
& 5.523
& 2.576
& 0.6297
& 0.4701
& 0.8150
& 24.1381
& 0.9325
\\

\cmidrule(lr){2-11}

&
\makecell[l]{Object\\Insertion}
&
Add-it~\citep{tewel2025addit}
&
{}
& 12.346
& 5.880
& 0.6162
& 0.4730
& 0.8234
& 23.8459
& 0.9101
\\

\cmidrule(lr){2-11}

&
\multicolumn{2}{l}{Ours (w/o context bbox)}
&
{}
& \underline{2.433}
& \textbf{1.516}
& \textbf{0.4839}
& \underline{0.4924}
& \underline{0.8493}
& 19.5459
& 0.5573
\\

&
\multicolumn{2}{l}{\textbf{Ours}}
&
{}
& \textbf{1.924}
& \underline{1.648}
& \underline{0.5913}
& \textbf{0.5400}
& \textbf{0.8626}
& \textbf{31.1420}
& \textbf{0.9820}
\\

\bottomrule
\end{tabular}
}
\end{table*}

\subsection{Detection Improvement}
To evaluate downstream utility, we compare SceneP2F with the unaugmented training baseline, Copy-Paste~\citep{Dwibedi_2017_ICCV}, and
X-Paste~\citep{pmlr-v202-zhao23f} on UAVDT~\citep{du2018unmannedaerialvehiclebenchmark},
where cars account for over 90\% of the instances. In separate bus- and truck-augmentation settings, each method augments 40\% of the training images by inserting two instances per image. The augmented images are added to the original training set, and a pretrained YOLO11-L is fine-tuned for 100 epochs with an image size of 1024, a batch size of 8.
As shown in Table~\ref{tab:rare-object-insertion}, SceneP2F improves bus/truck recall by 29.5\%/28.4\% and inserted-class $\mathrm{mAP}_{50}$ by 35.4\%/69.9\%, respectively, demonstrating the effectiveness of targeted augmentation for rare-class detection on UAVDT. Additional details and visualizations results are provided in Appendix~\ref{sec:appendix-detection}.

\begin{table*}
\caption{Downstream detection performance of rare-class augmentation methods on UAVDT
using YOLO11-L. In two separate settings, each method augments 40\% of the training images with either two buses or two trucks per image, and the resulting samples are added to the original training set. 
}
\label{tab:rare-object-insertion}
\centering
\small
\renewcommand{\arraystretch}{0.85}
\setlength{\tabcolsep}{5pt}

\begin{tabular}{@{}lllccccc@{}}
\toprule

\multirow{2}{*}{\bfseries Dataset}
&
\multirow{2}{*}{\makecell{\bfseries Inserted\\\bfseries Objects}}
&
\multirow{2}{*}{\bfseries Method}
&
\multicolumn{3}{c}{\bfseries Inserted Class}
&
\multicolumn{2}{c}{\makecell{\bfseries All Classes}}
\\

\cmidrule(lr){4-6}
\cmidrule(lr){7-8}

&
&
&
Recall
&
$\mathrm{mAP}_{50}$
&
$\mathrm{mAP}_{50\text{:}95}$
&
$\mathrm{mAP}_{50}$
&
$\mathrm{mAP}_{50\text{:}95}$
\\

\midrule

\multirow{8}{*}{\makecell{UAVDT\\~\citep{du2018unmannedaerialvehiclebenchmark}}}

& \multirow{4}{*}{2 buses}
& Original dataset
& 0.403 & 0.229 & 0.168 & 0.341 & 0.208 
\\

&
& Copy-and-Paste
& 0.408 & 0.268 & 0.190 & 0.358 & 0.211 
\\

&
& X-Paste
& 0.464 & 0.283 & 0.210 & 0.351 & 0.212 
\\

&
& \textbf{Ours (SceneP2F)}
& \textbf{0.522}
& \textbf{0.310}
& \textbf{0.226}
& \textbf{0.371}
& \textbf{0.224}
\\

\cmidrule(lr){2-8}

& \multirow{4}{*}{2 trucks}
& Original dataset
& 0.246 & 0.0718 & 0.045 & 0.341 & 0.208 
\\

&
& Copy-and-Paste
& 0.152 & 0.0659 & 0.0428 & 0.343 & 0.207 
\\

&
& X-Paste
& 0.202 & 0.0938 & 0.0590 & 0.347 & \textbf{0.209} 
\\

&
& \textbf{Ours (SceneP2F)}
& \textbf{0.316}
& \textbf{0.122}
& \textbf{0.0775}
& \textbf{0.351}
& 0.208
\\

\bottomrule
\end{tabular}
\end{table*}

\subsection{Ablations}
\begin{figure*}[t]
\centering
\captionsetup{font=footnotesize}

\begin{minipage}[t]{0.64\textwidth}
\caption{ScenePlan ablation. VRC denotes the mean valid-region coverage, Valid@0.9 the proportion of boxes with at least 90\% coverage, and ETC and TTC the exist--target and target--target collision rates.}
\label{tab:sceneplan-ablation}
\centering
\small
\renewcommand{\arraystretch}{0.9}
\setlength{\tabcolsep}{3.5pt}
\resizebox{\linewidth}{!}{
\begin{tabular}{@{}llccccc@{}}
\toprule
\bfseries Group
& \bfseries Method
& \bfseries VRC $\uparrow$
& \bfseries Valid@0.9 $\uparrow$
& \makecell{\bfseries 1-ETC $\uparrow$}
& \makecell{\bfseries 1-TTC $\uparrow$}
& \bfseries Average $\uparrow$
\\
\midrule
Full Model
& \textbf{Ours}
& \textbf{0.7671}
& \textbf{0.6339}
& 0.9057
& \textbf{0.9218}
& \textbf{0.8071}
\\
\midrule
\multirow{2}{*}{Size}
& SZ-xywh
& 0.7520
& 0.6050
& 0.8796
& 0.9150
& 0.7879
\\
& SZ-mean
& 0.7538
& 0.6163
& 0.9049
& 0.9188
& 0.7985
\\
\midrule
\multirow{3}{*}{Mask}
& MK-tok0
& 0.7489
& 0.6073
& 0.9074
& 0.8976
& 0.7903
\\
& MK-loss0
& 0.6895
& 0.5600
& 0.9021
& 0.9165
& 0.7670
\\
& MK-gate0
& 0.6887
& 0.5590
& \textbf{0.9076}
& 0.9119
& 0.7668
\\
\midrule
\multirow{2}{*}{Attention}
& ATT-L8
& 0.7543
& 0.6128
& 0.9028
& 0.8971
& 0.7918
\\
& ATT-Joint
& 0.6869
& 0.5429
& 0.8937
& 0.9129
& 0.7591
\\
\bottomrule
\end{tabular}
}
\end{minipage}
\hfill
\begin{minipage}[t]{0.35\textwidth}
\vspace{0pt}
\centering

\includegraphics[
    width=\linewidth,
    keepaspectratio
]{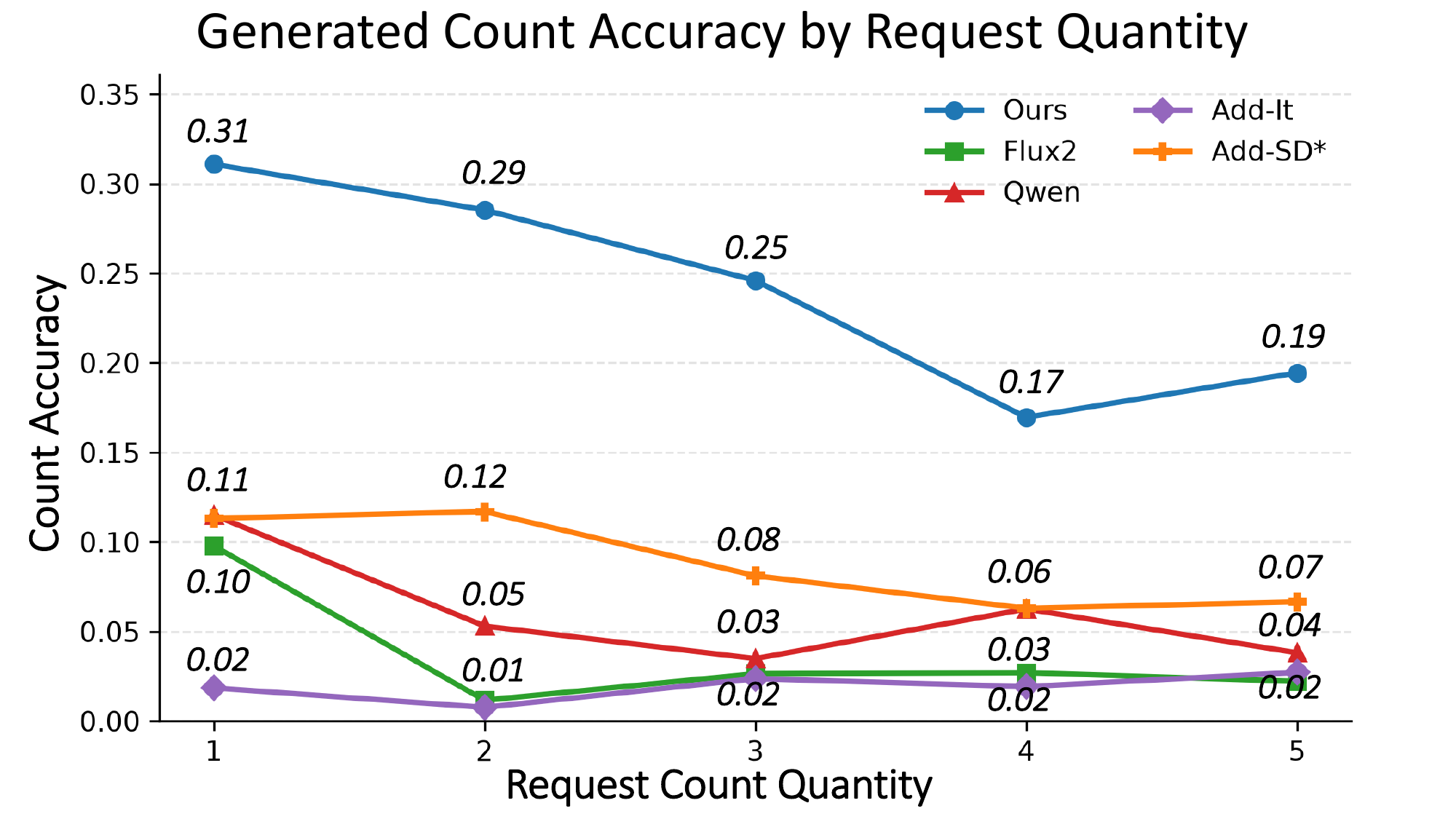}

\captionof{figure}{Mean count accuracy of the requested object count.
SceneP2F achieves the highest accuracy across request quantities. Add-SD* is averaged over UAVDT and VisDrone only.}
\label{fig:count_accuracy_graph}

\end{minipage}
\end{figure*}

\textbf{ScenePlan: Module Ablations}
We ablate ScenePlan using seven variants covering size prediction, region-mask conditioning, and attention design. SZ-xywh jointly predicts box coordinates and sizes, while SZ-mean uses mean sizes; tok0, loss0, gate0 remove the mask token, loss, and attention bias, respectively; and ATT-L8 and ATT-Joint use eight attention layers and joint self-attention throughout.
Tab.~\ref{tab:sceneplan-ablation} shows that the full model achieves the best overall placement validity and collision avoidance. For model selection and ablation only, we evaluate predicted layouts using valid regions obtained from SAM3 and four metrics: VRC (mean valid-region box coverage), Valid@0.9 (boxes with $\geq$90\% coverage), ETC (existing--target collisions), and TTC (target--target collisions).

\begin{figure}
    \centering
    \includegraphics[width=1.0\linewidth]{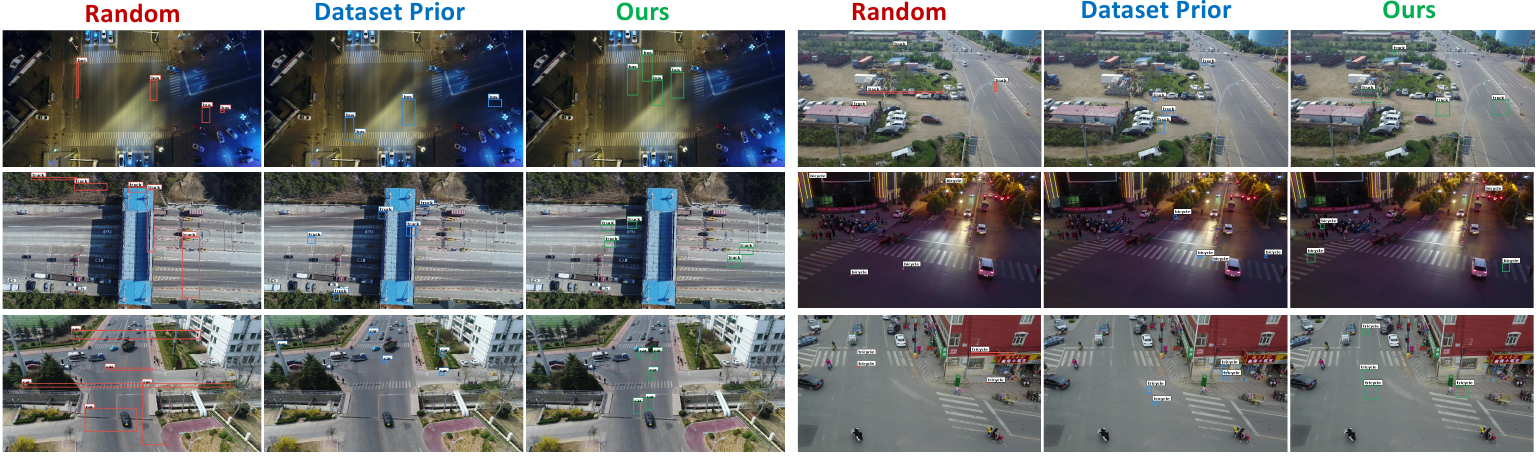}
    \caption{Qualitative comparison of Random, Dataset-Prior, and ScenePlan object placements on VisDrone and UAVDT, illustrating differences in predicted locations and scales.}
    \label{fig:comparison_box_planner}
\end{figure}

\textbf{ScenePlan: Box Prediction Comparison.}
Visualization in Fig.~\ref{fig:comparison_box_planner} compares ScenePlan with Random and Dataset-Prior. Random uniformly samples centers and draws box sizes from class-specific training ranges, whereas Dataset-Prior follows ODGEN~\citep{zhu2024odgen} and samples position, area, and aspect ratio from class-wise Gaussian priors. Across object categories and insertion counts, ScenePlan produces more plausible locations and scales, indicating better adaptation to scene.

\textbf{SceneAlign-Fill: Methods Comparison}
We compares SceneAlign-Fill with PowerPaint~\citep{zhuang2023task-powerpaint},
BrushNet~\citep{ju2024brushnet}, FLUX.1-Fill, and its LoRA-fine-tuned variant on 100 UAVDT removal-test images, with all specified boxes filled in a single pass. Results in Tab.~\ref{tab:image-editing-comparison} and Fig.~\ref{fig:fill-viz-comparison} show that SceneAlign-Fill performs better across all metrics, indicating improved category--count control, local appearance consistency, and semantic alignment rendering ability.

\begin{table*}[!htbp]
\centering
\captionsetup{font=footnotesize}
\begin{minipage}[t]{0.56\textwidth}
\vspace{0pt}
\centering
\captionof{table}{
Quantitative performance comparison of representative object-inpainting methods on the UAVDT removal test set in terms of request fidelity and within-box visual consistency.
}
\label{tab:image-editing-comparison}

\small
\renewcommand{\arraystretch}{0.9}
\setlength{\tabcolsep}{5pt}

\resizebox{\linewidth}{!}{
\begin{tabular}{@{}lccccc@{}}
\toprule

\bfseries Method
& \makecell{\bfseries mTCE $\downarrow$}
& \makecell{\bfseries mCCE $\downarrow$}
& \makecell{\bfseries LPIPS $\downarrow$}
& \makecell{\bfseries DINO$_{\mathrm{Sim}}$ $\uparrow$}
& \makecell{\bfseries CLIP-I $\uparrow$}
\\

\midrule

PowerPaint~\citep{zhuang2023task-powerpaint}
& 2.03
& 0.9403
& 0.6901
& 0.2598
& 0.7537
\\

BrushNet~\citep{ju2024brushnet}
& 1.68
& 0.7262
& 0.6551
& 0.2487
& 0.7346
\\

FLUX-Fill-Joint
& 2.12
& 0.6798
& 0.5495
& 0.3318
& 0.7872
\\

FLUX-Fill-Joint-FT
& 2.38
& 1.0731
& 0.5347
& 0.2940
& 0.7654
\\

\textbf{Ours (SceneAlign-Fill)}
& \textbf{0.77}
& \textbf{0.3388}
& \textbf{0.4229}
& \textbf{0.4128}
& \textbf{0.8252}
\\
\bottomrule
\end{tabular}
}
\end{minipage}
\hfill
\begin{minipage}[t]{0.43\textwidth}
\vspace{0pt}
\centering
\tiny
\captionof{table}{
SceneAlign-Fill Modules Ablation on
UAVDT. G0 for Geometry Token, ST0 for Style Token and TX-noR for regional attention.
}
\label{tab:uavdt-inside-edit-comparison}

\small
\renewcommand{\arraystretch}{0.9}
\setlength{\tabcolsep}{4pt}

\resizebox{\linewidth}{!}{
\begin{tabular}{@{}lccc@{}}
\toprule

\bfseries Method
& \bfseries LPIPS $\downarrow$
& \makecell{\bfseries DINO$_{\mathrm{Sim}}$ $\uparrow$}
& \bfseries CLIP-I $\uparrow$
\\

\midrule

G0
& 0.4305
& 0.3972
& 0.8195
\\

ST0
& 0.4335
& 0.3951
& 0.8183
\\

TX-noR
& 0.4339
& 0.4046
& 0.8201
\\

\textbf{Ours(SceneAlign-Fill)}
& \textbf{0.4229}
& \textbf{0.4128}
& \textbf{0.8252}
\\

\bottomrule
\end{tabular}
}

\end{minipage}
\end{table*}












\textbf{SceneAlign-Fill: Module Ablations.}
To validate the components of SceneAlign-Fill, we also conduct ablation experiments on 100 images from the UAVDT~\citep{du2018unmannedaerialvehiclebenchmark} removal test set. We separately remove the geometry token (G0) for target-position encoding, the style token (ST0) for color and style guidance, and the box-token regional attention (TX-noR) for region-specific semantic control. As shown in Tab.~\ref{tab:uavdt-inside-edit-comparison}, the full model outperforms all ablated variants, indicating that these components jointly improve the geometric, stylistic, and semantic consistency of inserted objects.












\section{Analysis and Conclusion}
\textbf{Limitation Analysis.}
Despite enabling scene-conditioned, category- and count-controlled multi-object insertion in aerial and road scenes, SceneP2F has several limitations. ScenePlan estimates object centers more reliably than box sizes, whose errors can propagate to SceneAlign-Fill and cause scale distortion, misalignment, or incomplete rendering. SceneAlign-Fill also struggles with extremely small objects and rare categories, sometimes producing fragmented or missing instances. Moreover, although SceneP2F improves category--count adherence, its absolute accuracy remains limited. Finally, qualitative observations suggest that GPT-Image-2.5-Pro can perform better on several aspects of the task. Since it was not included in the same controlled evaluation as the open-source baselines, we do not claim superiority over it; a unified comparison is left for future work.

\textbf{Conclusion.}
We presented SceneP2F, a two-stage framework for scene-conditioned, category- and count-controlled multi-object insertion. By connecting \textsc{ScenePlan} and \textsc{SceneAlign-Fill} through explicit class-labeled layouts, the framework coordinates scene-compatible placement with annotation-aligned rendering while promoting source-content preservation. 
Experiments on challenging aerial and road scenes demonstrate improved request and scene fidelity, while targeted augmentation on UAVDT~\citep{du2018unmannedaerialvehiclebenchmark} improves detection of underrepresented categories. 
These findings highlight the importance of maintaining instance-level consistency across requests, layouts, and generated objects to turn visually plausible insertions into useful annotated training data.


\subsection*{AI use statement}
We used generative AI tools for language polishing, improving manuscript readability, discussing research methodology and experimental design, and assisting with minor code utilities and debugging. All AI-assisted outputs were reviewed and verified by the authors. The authors take full responsibility for all contents, claims, and artifacts in this work.

\subsection*{Ethics statement}
This work investigates controllable object insertion as a means of reducing the cost of constructing and annotating object-detection datasets, particularly for underrepresented categories and challenging conditions. Synthetic data may inherit biases or artifacts from source datasets and pretrained models, and their associated annotations may contain errors. Models trained with such data should therefore be evaluated on real-world data and should not be deployed in safety-critical settings without task-specific validation.

\subsection*{Reproducibility Statement}
The main paper and appendix describe the proposed method, data-processing pipeline, training and inference settings, evaluation protocol, and downstream detection experiments. To support independent verification,  we intend to make the associated implementation artifacts available after publication where permitted by applicable dataset and model licenses.

\bibliography{iclr2027_conference}
\bibliographystyle{iclr2027_conference}

\appendix
\section{Appendix: removal dataset preparation}
\subsection{removal methods comparison}
\label{removal methods comparion}
\begin{figure}[!t]
    \centering
    \includegraphics[width=1.0\linewidth]{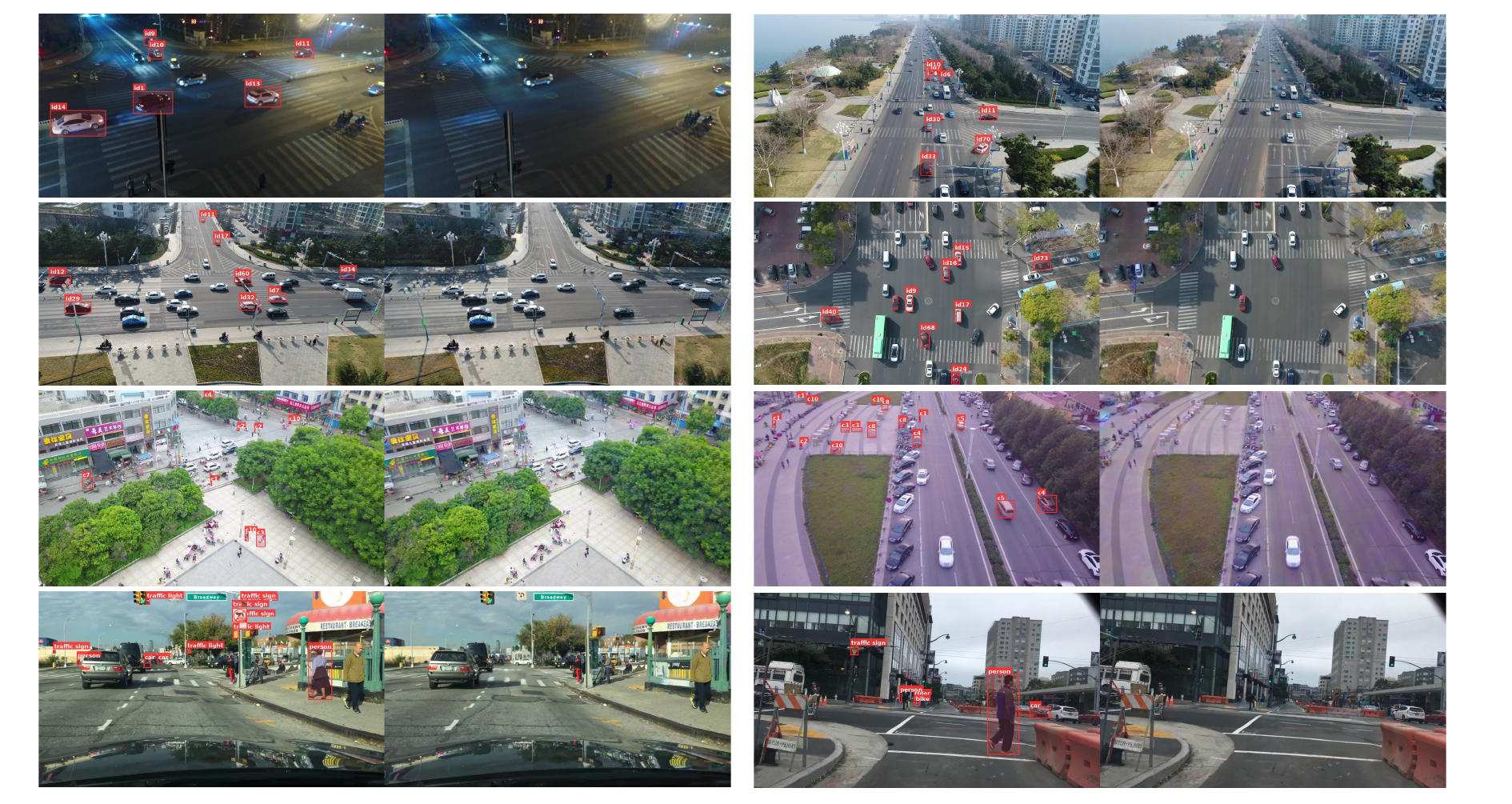}
    \caption{Visualization of Removal Dataset Processed images. The left figure is the original image with target removed bbox and corresponding class. while the right is processed image. }
    \label{fig:removal-datasets-viz}
\end{figure}

We compared and analysis four different removal methods.
First, LaMa employs a single-stage inpainting network with fast Fourier convolutions and a high-receptive-field perceptual loss, providing efficient reconstruction of large masks and strong generalization to high-resolution images, but without explicitly modeling object-associated effects such as shadows and reflections~\citep{suvorov2022resolution}. DesignEdit instead adopts a training-free latent-diffusion framework that decomposes an image into object and background layers, performs background completion through key-masked self-attention, and suppresses artifacts during latent fusion; it supports diverse spatial edits but is less specialized for precise object-effect removal~\citep{jia2024designedit}. OmniPaint jointly formulates insertion and removal using a pretrained diffusion prior and a progressive training strategy that combines paired supervision with large-scale unpaired refinement through CycleFlow, yielding improved foreground elimination and scene consistency at the cost of iterative diffusion inference~\citep{yu2025omnipaint}. ObjectClear~\citep{zhao2026objectclear} explicitly decouples foreground removal from background reconstruction through adaptive target-aware attention and further protects unmasked content using attention-guided fusion~\citep{zhao2026objectclear}. Consequently, LaMa offers the best efficiency and resolution robustness, DesignEdit provides training-free editing flexibility, OmniPaint balances insertion and removal within a unified model, whereas ObjectClear achieves the strongest reported object-effect removal and background fidelity, particularly in scenes containing shadows, reflections, or multiple interacting objects. The final visualization results are shown in Fig.~\ref{fig:removal-datasets-viz}.

\subsection{implement details}
\label{preparation details}

\paragraph{UAVDT~\citep{du2018unmannedaerialvehiclebenchmark}.}
We split the 50 UAVDT sequences into training and test sets at a 4:1 ratio. The test set comprised the following 10 sequences: m0101, m0202, m0207, m0208, m0209, m0210, m0704, m0901, m1102, and m1202. We use UAVDT frames and GT bbox annotations to construct the datasets. Cars, trucks, and buses are eligible for removal if their bounding boxes contain at least $8^2$ pixels, do not touch image boundaries, and have zero overlap with other instances; all ineligible instances are retained as context. We sample the removal count $k$ from $\{1,2,3,4,6,8,12,16,18\}$ with probabilities $\{0.1,0.1,0.1,0.1,0.1,0.2,0.1,0.1,0.1\}$ and set
\[
k \leftarrow \min\left(k,18,\left\lfloor 0.5n_{\mathrm{valid}}\right\rfloor\right).
\]

\paragraph{VisDrone~\citep{zhu2021detection}.}
We process the VisDrone~\citep{zhu2021detection} validation split, with support for the train and test-dev splits. Categories 1--10 are removable, whereas ignored instances, non-positive-score boxes, and categories 0 and 11 are excluded from target selection but retained as context when applicable. Candidates must contain at least $64$ pixels, be non-truncated, and have zero overlap with other boxes. We sample $k$ from $\{1,2,3,4,6,8,12,16,18,24\}$ with probabilities $\{0.09,0.09,0.09,0.09,0.09,0.18,0.09,0.09,0.09,0.10\}$ and clip it by
\[
k \leftarrow
\min\left(
k,\,
\begin{cases}
24, & n_{\mathrm{target}}\geq 38,\\
18, & \text{otherwise},
\end{cases}
\left\lfloor 0.5n_{\mathrm{target}}\right\rfloor
\right).
\]
Targets are selected using dataset-level category deficits and within-frame category round-robin sampling to improve class balance.

\paragraph{BDD100K~\citep{yu2020bdd100kdiversedrivingdataset}.} Only rectangular instances from nine classes---person, rider, bike, car, truck, bus, motor, traffic light, and traffic sign---are eligible for removal; lanes, drivable areas, trains, non-rectangular objects, and out-of-vocabulary categories are used only as context. Candidates must contain at least $64$ pixels, be non-truncated and boundary-separated, and have a maximum IoU below $0.2$ with other training instances. We sample $k$ from $\{1,2,3,4,6,8,10,12\}$ with probabilities $\{0.12,0.12,0.12,0.12,0.14,0.16,0.10,0.12\}$ and set
\[
k \leftarrow \min\left(k,12,
\left\lfloor 0.5n_{\mathrm{target}}\right\rfloor,
n_{\mathrm{target}}\right).
\]
To mitigate class imbalance, target sampling allocates $65\%$ of samples to major classes (car, traffic sign, traffic light, person, and truck) and $35\%$ to tail classes (bus, bike, rider, and motor), followed by category balancing within each group.

\section{Appendix: Plausible region mask construction}
\label{Valid region mask construction}
\begin{figure}
    \centering
    \includegraphics[width=1.0\linewidth]{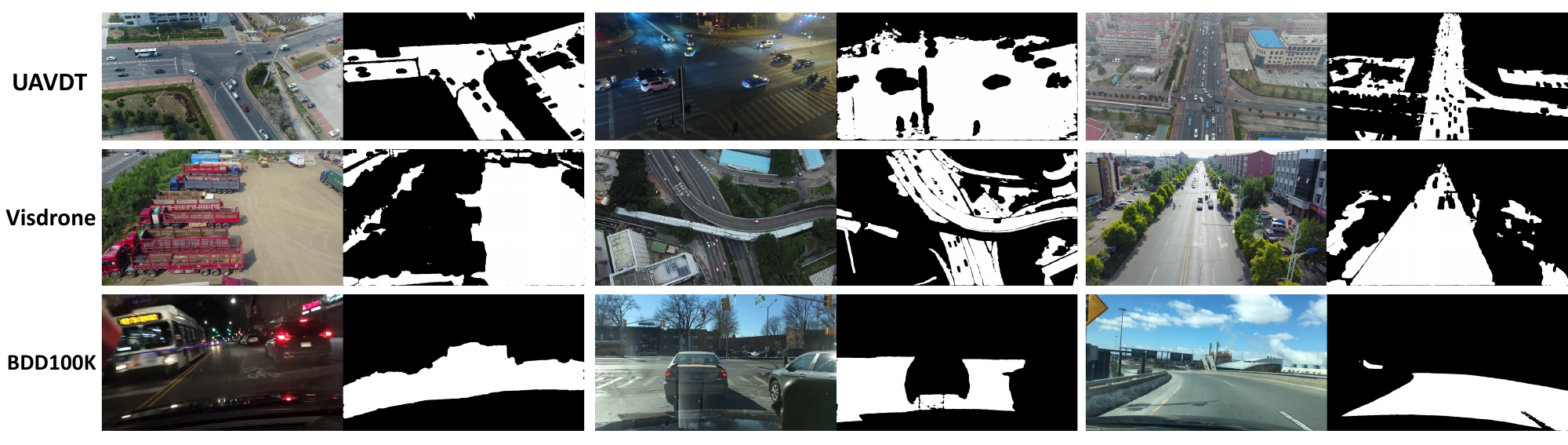}
    \caption{Visualization of plausible region mask among three datasets generated by SAM3. The black and white image on the right of each RGB image is the binary map of the result segmentation.}
    \label{fig:mask_visualization}
\end{figure}

All datasets use SAM3~\citep{carion2025sam3segmentconcepts} masks with a four-pixel dilation, no expansion of protected regions, FP16 inference, and quality filtering with an edge-density threshold of $0.05$ and an image-difference threshold of $15.0$. UAVDT~\citep{du2018unmannedaerialvehiclebenchmark} and VisDrone~\citep{zhu2021detection} use at most five batches per sample and enable a second removal pass, with at most four and five boxes per group, respectively; BDD100K\citep{yu2020bdd100kdiversedrivingdataset} uses four batches, three boxes per group, and a single pass. We present visualization of image and corresponding binary mask in Fig.~\ref{fig:mask_visualization}.

\textbf{VisDrone.}
We use the SAM3~\citep{carion2025sam3segmentconcepts} prompt configuration to separately segment roadway, pedestrian/footway, and open-ground support regions. Generic terms such as \emph{pavement}, \emph{ground}, \emph{street}, and \emph{concrete} are omitted to avoid merging roads with sidewalks. Four-wheel vehicles are constrained to roadway masks, light vehicles to roadway and footway masks excluding pedestrian-open regions, and pedestrians to the union of all support channels.

\textbf{BDD100K.}
We generate a single-channel road-support mask using the prompt set:
\{\emph{road}, \emph{road surface}, \emph{drivable road}, \emph{traffic lane}, \emph{street}, \emph{intersection}, \emph{parking lot}, \emph{road shoulder}, \emph{driveway}, \emph{crosswalk on road}\}.
Masks are generated at resolution $1008$, thresholded at $0.4$; traffic signs and lights are excluded from ground-based placement.

\textbf{UAVDT.}
For each UAVDT frame, we generate a vehicle-support mask using the SAM3 prompts \{\emph{roads}, \emph{street}, \emph{parking areas}\}. Inference uses a confidence threshold of $0.4$, skips completed outputs, and stores one binary mask per frame.


\section{Appendix: Details of ScenePlan}
\subsection{Pre-Trained Size Predictor}
\label{sec:appendix-size-predictor}

\textbf{Ablations.} 
As shown in Tab.\ref{tab:size-predictor-ablation}, we evaluate the ScenePlan size predictor on 100 images from the UAVDT~\citep{du2018unmannedaerialvehiclebenchmark}
removal test set, sampling 10 images from each sequence. For each removed
object, the center coordinates $(x,y)$ of its ground-truth bounding box
are provided as input, and the predictor estimates its width and height
$(w,h)$. We compute class-balanced L1 errors between the predicted and
ground-truth sizes in the log-area and aspect-ratio spaces. As shown in
Table~\ref{tab:size-predictor-ablation}, the complete model achieves an
average error of $0.3691$. Removing the entire global-feature branch
(GF0) increases the error to $0.3895$, while removing the box-distance
cue (GF1) or the median area of neighboring context boxes (GF2) results
in errors of $0.3777$ and $0.3825$, respectively. The relative-feature
branch has a greater effect: removing the entire branch (RF0) increases
the error to $0.4350$, while removing the context-class embeddings (RF1)
or relative-position features (RF2) yields $0.3775$ and $0.4012$,
respectively. These results indicate that relative geometry is
particularly important for scene-conditioned size estimation. For the
prediction head, our two-layer architecture
($352\rightarrow256\rightarrow256\rightarrow2$) outperforms the
shallower, deeper, narrower, and wider alternatives, suggesting that
moderate model capacity provides the best trade-off for this task.
Finally, reducing the learning rate from $10^{-3}$ to
$3\times10^{-4}$ produces a comparable error of $0.3713$, whereas
increasing it to $3\times10^{-3}$ degrades the error to $0.4008$,
supporting the use of $10^{-3}$ as the default learning rate.

\begin{table*}[t]
\centering
\caption{
Ablation study of the ScenePlan size predictor. We evaluate alternative
global-feature and relative-feature designs, MLP head architectures, and
training settings using class-balanced log-area and aspect-ratio L1
errors. The average is the arithmetic mean of the two errors. }
\label{tab:size-predictor-ablation}

\small
\renewcommand{\arraystretch}{1.05}
\setlength{\tabcolsep}{5pt}

\begin{tabular}{@{}llcccl@{}}
\toprule
\bfseries Group
& \bfseries ID
& \makecell{\bfseries Class-balanced\\\bfseries log-area L1 $\downarrow$}
& \makecell{\bfseries Class-balanced\\\bfseries aspect-ratio L1 $\downarrow$}
& \bfseries Average $\downarrow$
& \bfseries Description \\
\midrule

\multicolumn{2}{l}{\textbf{Ours}}
& 0.4651
& 0.2730
& 0.3691
& Selected configuration \\

\midrule
\multirow{3}{*}{Global feature}
& GF0
& 0.5057
& 0.2732
& 0.3895
& All global features \\

& GF1
& 0.4621
& 0.2933
& 0.3777
& Distance feature \\

& GF2
& 0.4774
& 0.2876
& 0.3825
& Median-area feature \\

\midrule
\multirow{3}{*}{Relative feature}
& RF0
& 0.5149
& 0.3550
& 0.4350
& All relative features \\

& RF1
& 0.4834
& 0.2715
& 0.3775
& Class embedding \\

& RF2
& 0.4853
& 0.3170
& 0.4012
& Relative geometry \\

\midrule
\multirow{4}{*}{MLP head}
& HM-d1
& 0.4759
& 0.3165
& 0.3962
& One hidden layer \\

& HM-d4
& 0.4877
& 0.3819
& 0.4348
& Four hidden layers \\

& HM-w128
& 0.4913
& 0.2719
& 0.3816
& Hidden width of 128 \\

& HM-w512
& 0.5014
& 0.5018
& 0.5016
& Hidden width of 512 \\

\midrule
\multirow{2}{*}{Training settings}
& TS-lr$\downarrow$
& 0.4626
& 0.2799
& 0.3713
& $\mathrm{lr}=3\times10^{-4}$ \\

& TS-lr$\uparrow$
& 0.4612
& 0.3403
& 0.4008
& $\mathrm{lr}=3\times10^{-3}$ \\

\bottomrule
\end{tabular}
\end{table*}

\subsection{Box Planner Post-Process}
\label{sec:appendix-post-process}
The module repairs each box onto a feasible set defined by image bounds, scene-valid masks, and existing object layouts, while preserving the planner's semantic class assignment and insertion count. When a predicted box violates feasibility, we repair it using constrained translations and, when necessary, mild scale or
size re-estimation, yielding final boxes that are both spatially valid and less likely to collide with observed objects.

For VisDrone\citep{zhu2021detection} and UAVDT\citep{du2018unmannedaerialvehiclebenchmark} aerial datasets, we first clamps each raw box to image bounds and size limits, and then applies an iterative procedure against the union plausible region mask mentioned above, context boxes, and other target boxes; this procedure allows bounded mask shifts, context/target separation shifts, and small scaling when overlap or invalid-mask coverage remains. For BDD100K\citep{yu2020bdd100kdiversedrivingdataset}, we use a rule-based repair tailored to street-view geometry: if a predicted insertion overlaps an existing
vehicle, we translate it only along the horizontal axis until the collision is removed, keeping its predicted size fixed;
otherwise, we snap its center vertically into the road/ground valid mask, re-estimate its size with the frozen size predictor, and optionally match its area to the nearest upper 
vehicle in the same horizontal direction. Thus, both pipelines enforce validity after layout prediction and before inpainting, but BDD100K\citep{yu2020bdd100kdiversedrivingdataset} uses dataset-specific street-scene heuristics with size predictor.

\section{Appendix: Details of SceneAlign-Fill}
\subsection{Model Details}
\label{sec:appendix_SceneAlign-Fill}
To equip FLUX.1-Fill with
instance-level layout control and source-scene appearance awareness, we
introduce three complementary conditioning signals: dense geometry
features, region-level box tokens, and a context-derived appearance
code.

\textbf{Dense geometry conditioning.}
For each packed image token at normalized coordinate
$\boldsymbol{\xi}_n$, we identify its nearest valid insertion box
$\hat{b}_i^*$ and encode the token's relative position and box scale as
\begin{equation}
\mathbf g_n =
\left[
(\boldsymbol{\xi}_n-\boldsymbol{\mu}_{\hat{b}_i^*})^\top,
\log\max(w_{\hat{b}_i^*},\delta),
\log\max(h_{\hat{b}_i^*},\delta)
\right]^\top .
\end{equation}
Here, $i=i(n)$ indexes the nearest valid box,
$\boldsymbol{\mu}_{\hat{b}_i^*}$, $w_{\hat{b}_i^*}$, and
$h_{\hat{b}_i^*}$ denote its normalized center, width, and height,
respectively, and $\delta>0$ ensures numerical stability.
The resulting dense geometry field provides each image token with explicit
spatial information relative to the planned layout.

\textbf{Box-token regional attention.}
For each valid insertion box, we combine its normalized geometry with a
frozen CLIP class embedding to construct a box token.
These tokens are appended to the global text-conditioning sequence. In
the final four FLUX blocks, an image token can attend to a box token
only when its spatial location lies inside the corresponding box,
whereas the global text tokens remain accessible to all image regions.
This region-restricted attention associates category semantics with
their designated insertion regions and reduces interference between
instances.

\textbf{Context-derived appearance conditioning.}
To align the inserted objects with the source scene, we extract color
statistics and ROI-pooled VAE features exclusively from existing
context objects. The two descriptors are projected, fused through layer
normalization, and modulated by a learned scalar gate to produce a
global appearance code $\tau_s$. When no valid context
object is available, a learned null-style embedding is used instead.
The appearance is then broadcast to all image-token positions.

Finally, the native FLUX.1-Fill condition is augmented with the dense
geometry and appearance features:
\begin{equation}
    C_p=
    \left[
        z_{\mathrm{cond},p};
        \tau_{m_p};
        \tau_g;
        \tau_s
    \right],
    \label{eq:compact_condition}
\end{equation}
Together, these conditions provide instance-level geometric control,
region-specific category guidance, and scene-consistent appearance
conditioning for multi-object insertion.

\paragraph{Training and inference.}
\label{sec:SceneAlign-Fill_training_and_inference}

We train SceneAlign-Fill using rectified flow matching. Given the target-image
latent $z_0$, we sample a timestep $t\sim\mathcal{U}(0,1)$ and Gaussian noise
$\epsilon\sim\mathcal{N}(0,\mathbf I)$ of the same dimensionality, and construct
\begin{equation}
z_t=(1-t)z_0+t\epsilon,
\qquad
v^\star=\epsilon-z_0,
\end{equation}
where $z_t$ is the interpolated noisy latent and $v^\star$ is its target
velocity. The model predicts the conditional velocity field as
\begin{equation}
\hat{v}_{\theta}
=
F_{\theta}
\left(
z_t,
C,
\mathcal{U}_{\mathrm{box}},
y,
t
\right),
\end{equation}
where $F_{\theta}$ is the conditional velocity predictor,
$C=\{C_p\}_{p=1}^{N}$ denotes the image-side conditioning tokens,
$\mathcal{U}_{\mathrm{box}}
=\{u_i\}_{i=1}^{N_{\mathrm{tgt}}}$ is the set of box tokens, and $y$ is the
global text condition.

To emphasize annotation-aligned synthesis while preserving the source scene,
we use the spatially weighted flow-matching objective
\begin{equation}
\mathcal{L}_{\mathrm{FM}}
=
\mathbb{E}_{t,\epsilon}
\left[
\frac{1}{N}
\sum_{p=1}^{N}
\omega_p
\left\|
\hat{v}_{\theta,p}-v^\star_p
\right\|_2^2
\right],
\qquad
\omega_p=
\begin{cases}
\lambda_{\mathrm{box}}, &
p\in\displaystyle\bigcup_i b_i,\\
1, & \mathrm{otherwise},
\end{cases}
\label{eq:compact_objective}
\end{equation}
where $N$ is the number of packed latent tokens, $p$ indexes their spatial
positions, and $b_i$ denotes the region covered by the $i$-th target box.
The weight $\lambda_{\mathrm{box}}>1$ emphasizes velocity prediction within
the annotated regions, while the unit weight elsewhere penalizes unintended
changes to the preserved scene.

We freeze the VAE, text encoders, and pretrained FLUX.1-Fill backbone weights,
and optimize only the newly introduced conditioning modules, expanded input
projection, and LoRA adapters. At inference time, predicted or ground-truth
layouts are converted into the same image-side conditions and box tokens. The
conditional rectified-flow trajectory is then integrated from Gaussian noise,
and the resulting latent is decoded by the frozen VAE to produce the output
image.

\subsection{Fill Methods Comparison}
\label{sec:appendix-fill-viz-comparison}
\begin{figure}
    \centering
    \includegraphics[width=1.0\linewidth]{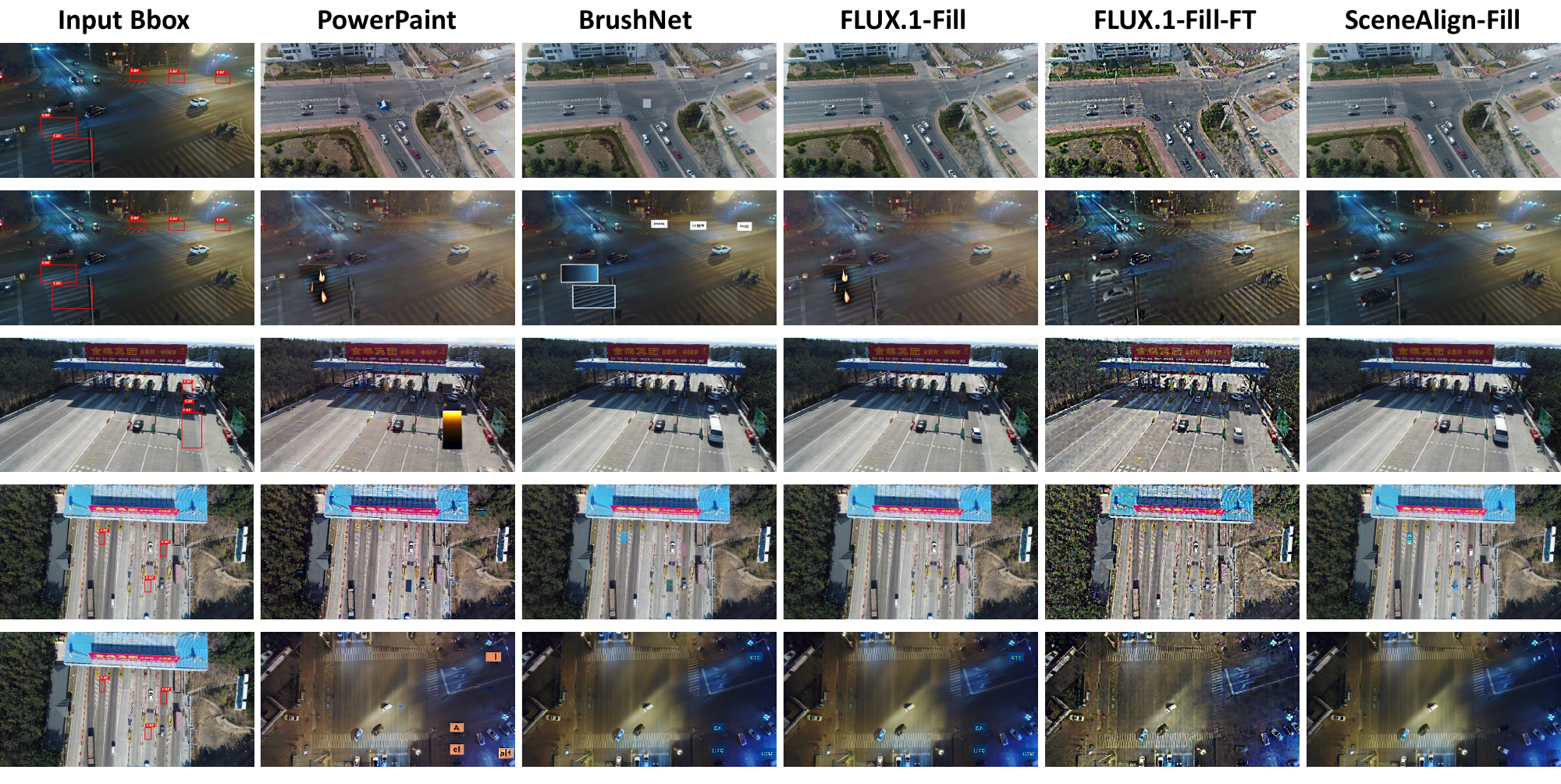}
    \caption{Visual Comparison of different Image Inpainting Methods with SceneAlign-Fill.}
    \label{fig:fill-viz-comparison}
\end{figure}

\section{Model training and Evaluation}
\label{sec:appendix-training-infernce}

We train a separate model for each dataset using the same two-stage
pipeline. We first train a size predictor \(G_{\phi}\) to estimate box
width and height from a target center and its scene context. After
freezing \(G_{\phi}\), we train \textsc{ScenePlan} to jointly predict
the centers of the requested instances. The predicted centers and sizes
are combined into complete boxes and adjusted by the post-processing procedure. The resulting layout is then passed to
\textsc{SceneAlign-Fill} for image synthesis.

\subsection{Training Detail Settings}
\label{sec:appendix_SceneP2F_training_and_inference}
\paragraph{Pre-Trained Size predictor.}
All size predictors are trained independently for \(50\) epochs using
AdamW with weight decay \(10^{-4}\) and an effective batch size of
\(64\). They are subsequently frozen and incorporated into
\textsc{ScenePlan} as geometric conditions. UAVDT and BDD100K use a
learning rate of \(10^{-3}\), \(500\) warm-up steps, and gradient
clipping at \(1.0\). For Visdrone, We use a \(352\rightarrow512\rightarrow512\rightarrow2\) MLP with a
learning rate of \(3\times10^{-3}\). The training, validation, and
test-dev removal splits are used for training, model selection, and
final evaluation, respectively.For UAVDT, the predictor directly regresses log width and height. To mitigate class
imbalance, the losses for cars, trucks, and buses are weighted by
\(0.4\), \(0.4\), and \(0.2\), respectively. For BDD100K, we use a \(352\rightarrow256\rightarrow256\rightarrow2\) MLP trained on
the training removal set, with the validation removal set used for model
selection. DINOv2-Large features are limited to \(256\) tokens and
\(128\) context boxes; center and validation-box jitter are set to
\(0.03\) and \(0.05\), respectively, and the loss is macro-averaged
across classes.

\paragraph{\textsc{ScenePlan} training.}
All planner variants use DINOv2-Large as the image encoder. The planner contains four Transformer layers with a hidden dimension of \(512\) and eight attention heads. It supports at most \(18\) target queries and \(128\) context boxes, and is trained using a \(100\)-step diffusion process with velocity prediction.

For UAVDT, we use a learning rate of
\(5\times10^{-5}\), weight decay \(0.01\), and a batch size of \(64\);
the selected checkpoint is obtained at step \(20{,}000\). Both aerial
planners use precomputed SAM3 support-region masks. The masks are
encoded into valid-region summary tokens and used to softly modulate
image attention. A center-validity loss with weight \(0.5\) further
encourages planned boxes to lie in plausible regions. For VisDrone2019, we train the planner for \(20{,}000\) optimization
steps using AdamW with a learning rate of \(2\times10^{-5}\) and an
effective batch size of \(64\). 

The BDD100K planner is trained without scene-validity masks,
mask-guided attention, or the center-validity loss. Its active insertion
categories are \emph{person}, \emph{rider}, \emph{bike}, \emph{car},
\emph{truck}, \emph{bus}, \emph{motor}, \emph{traffic light}, and
\emph{traffic sign}; \emph{lane} and \emph{drivable area} are excluded.
Although the model reserves \(11\) class slots, only these nine object
categories are used as insertion targets.

\paragraph{\textsc{SceneAlign-Fill} training.}
All dataset-specific models are initialized from FLUX.1-Fill-dev, with
the pretrained VAE, text encoders, and base-model parameters frozen. We
optimize only the LoRA adapters, expanded image-input projection, and
new conditioning modules at a resolution of \(768\), using a
per-device batch size of \(1\), four gradient-accumulation steps,
\texttt{bfloat16} precision, and seed \(2024\). Training is limited to
\(50{,}000\) steps with an early-stopping patience of \(5{,}000\)
steps.
For UAVDT, we use learning rates of \(5\times10^{-5}\) and \(10^{-4}\) for the conditioning modules and LoRA adapters, respectively, and select the
checkpoint at step \(11{,}000\). Class-balanced sampling is enabled,
with insertion classes weighted by the square root of inverse class
frequency. For Visdrone, the conditioning modules and LoRA adapters use learning rates of
\(10^{-5}\) and \(3\times10^{-5}\), respectively. The selected
checkpoint is obtained at step \(4{,}000\). For BDD100K, the conditioning modules and LoRA adapters use learning rates of
\(5\times10^{-5}\) and \(10^{-4}\), respectively; the LoRA adapters
have rank \(16\), scaling factor \(16\), and dropout \(0.05\). 

\subsection{Details of Baseline Model Fine-tuned}
\label{sec:appendix-baseline model finetune}
For a controlled comparison, we adapt Add-it  and Qwen separately to UAVDT and VisDrone using the same object-removed original image pairs and insertion prompt template as our method. Training uses 32,297 pairs from the designated 40 UAVDT training sequences and 6,404 pairs from the official VisDrone training split. We freeze the FLUX.1-dev backbone and train dataset-specific LoRA adapters on its attention projections, with rank and scaling factor both set to 16. Images are resized while preserving their aspect ratios and padded to $1024\times1024$. Each adapter is trained for 2,000 optimization steps using AdamW, a learning rate of $5\times10^{-5}$, weight decay of 0.01, 100 warm-up steps, and an effective batch size of 4, with BF16 precision on a single NVIDIA H200 GPU. We use an object-region-weighted flow-matching loss ($\lambda_{\mathrm{obj}}=10$) together with an L1 background-preservation loss ($\lambda_{\mathrm{bg}}=1$), deriving supervision masks from the removed-object bounding boxes. The final-step adapters are used for evaluation without test-set-based checkpoint selection.

\subsection{Appendix: Evaluation Methods}
\label{sec:appendix-eval-details}
\subsubsection{Dataset-Specific Insertion Test Schedule.}

All three schedules combine equal quotas, category balancing, and GT-admissible or scene-consistency constraints to produce reproducible and auditable insertion requests. By reducing the influence of natural category frequencies and scene densities, these schedules more directly evaluate the planner's controllability over object count, category composition, placement, and scale.

\textbf{UAVDT.}
For UAVDT, test requests are constructed at the scene level using three vehicle classes: \texttt{car}, \texttt{truck}, and \texttt{bus}. The Common Track assigns equal quotas to $N\in\{1,\ldots,5\}$. For $N=1$, only single-class requests are used; for $N\geq2$, the schedule contains $60\%$ single-class and $40\%$ mixed-class requests. The three classes are balanced symmetrically across both request types, preventing the naturally dominant \texttt{car} class from suppressing the evaluation of \texttt{truck} and \texttt{bus}.

\textbf{Visdrone.}
We evaluate Visdrone on the official test-dev split, covering 10 active categories of pedestrians, vehicles, and lightweight transportation objects. The Common Track assigns equal quotas to $N\in\{1,\ldots,5\}$. Each image may request only categories present in its original ground-truth annotations, which we refer to as the \emph{GT-admissibility} constraint. Subject to this constraint, we maintain global category balance and a target ratio of $60\%$ single-class to $40\%$ mixed-class requests whenever feasible. This design accommodates the dataset's diverse, long-tailed categories and varying scene densities while avoiding semantically implausible insertions.

\textbf{BDD100K.}
For BDD100K, we sample a fixed set of 1,000 independent images from the official validation split and allocate 200 images to each $N\in\{1,\ldots,5\}$. We use the same nine detection classes as in training: \texttt{person}, \texttt{rider}, \texttt{bike}, \texttt{car}, \texttt{truck}, \texttt{bus}, \texttt{motor}, \texttt{traffic light}, and \texttt{traffic sign}. Non-generative categories, such as \texttt{train}, \texttt{drivable area}, and \texttt{lane}, are explicitly excluded. Requests are restricted to categories visible in the original ground truth. For $N=1$, admissible categories are sampled uniformly; for $N\geq2$, we target a $60\%$ single-class and $40\%$ mixed-class ratio and apply global rebalancing to increase the exposure of long-tailed.

\subsubsection{Detect Added Objects}
\label{sec:appendix-detector-gd}
\begin{figure}
    \centering
    \includegraphics[width=1.0\linewidth]{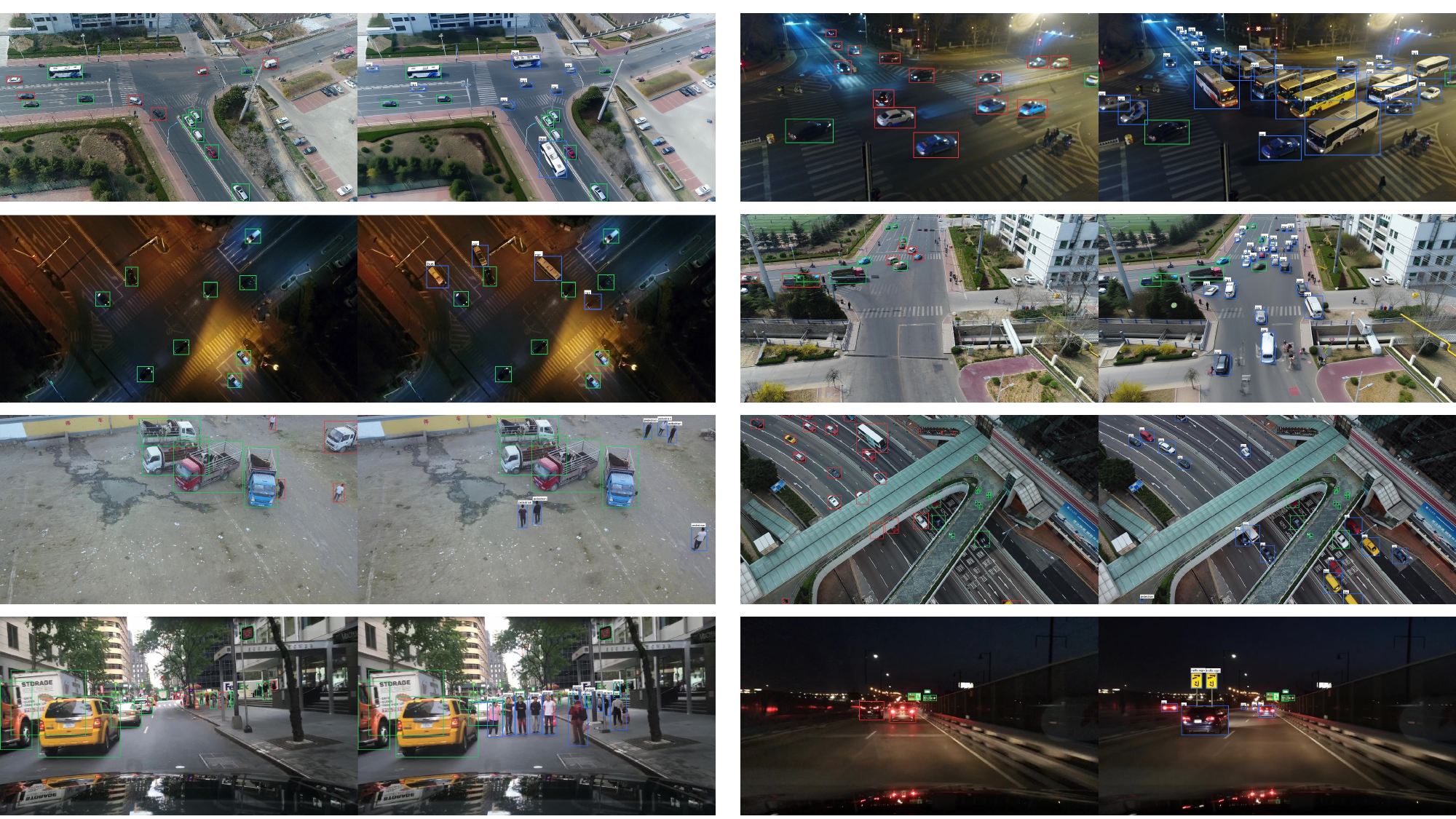}
    \caption{Visualization of inserted object detected by GroundingDINO. The figure in the left in original image and right is the generated. Green, red, and blue boxes indicate matched, unpreserved, and newly inserted objects, respectively; inserted-object metrics are computed on the blue boxes.}
    \label{fig:gd-vis}
\end{figure}

We fine-tune GroundingDINO with a Swin-B backbone, using a batch size of 2 for 12 epochs across three datasets. To identify inserted objects, we apply the trained detector to both the source and generated images, using the concatenated dataset class names as the caption. Same-class detections with an IoU of at least 0.50 are greedily matched, and unmatched detections in the generated image are treated as inserted-object boxes. We use score thresholds of 0.25, followed by class-wise NMS with an IoU threshold of 0.70. As visualization shown in Fig.\ref{fig:gd-vis}, green boxes denote matched objects, red boxes denote unmatched source objects treated as unpreserved, and blue boxes denote newly inserted objects. All subsequent inserted-object metrics are computed using the blue boxes. We present the accuracy of our trained GroundingDINO in Tab.\ref{tab:groundingdino_eval}.

\begin{table}[t]
\centering
\caption{Evaluation results of GroundingDINO on BDD100K, UAVDT and Visdrone test dataset}
\label{tab:groundingdino_eval}
\begin{tabular}{lccc}
\hline
Dataset & Recall & mAP50 & mAP50--95 \\
\hline
BDD100K~\citep{yu2020bdd100kdiversedrivingdataset} & 94.91 & 65.54 & 38.04 \\
UAVDT~\citep{du2018unmannedaerialvehiclebenchmark} & 96.64 & 91.71 & 70.75 \\
VisDrone~\citep{zhu2021detection} & 78.88 & 48.29 & 28.59 \\
\hline
\end{tabular}

\end{table}

\subsubsection{Metrics Calculation Details}
\label{sec:appendix-metrics}
\paragraph{Request Fidelity}
We calculate Mean Total Count Error (mTCE) and Mean Class Count Error (mCCE): 
\[
\mathrm{mTCE}=\frac{1}{N}\sum_{i=1}^{N}\left|\hat{T}_i-T_i\right|,\qquad \mathrm{mCCE}=\frac{1}{N}\sum_{i=1}^{N}\frac{\sum_{c\in\mathcal{C}}\left|\hat{n}_{i,c}-n_{i,c}\right|}{\sum_{c\in\mathcal{C}}n_{i,c}} .
\]
Here, $T_i$ and $\hat{T}_i$ denote the requested and detected total counts, while $n_{i,c}$ and $\hat{n}_{i,c}$ denote their category-wise counterparts; lower values indicate better request fidelity.

\section{Datasets Augmentation for Detection}
\label{sec:appendix-detection}
\begin{figure}
    \centering
    \includegraphics[width=1.0\linewidth]{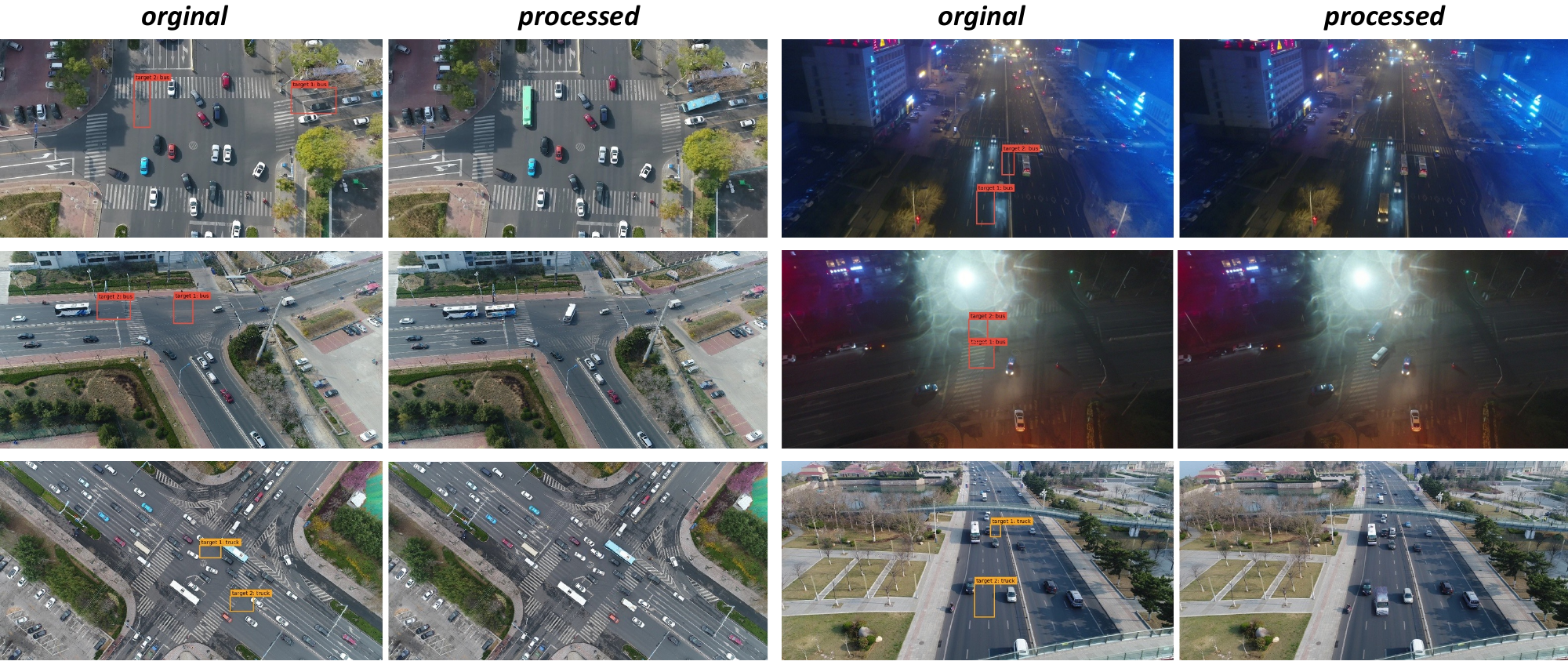}
    \caption{Qualitative results for the two-bus and two-truck insertion settings on UAVDT.}
    \label{fig:uavdt-aug}
\end{figure}

For Copy--Paste, we build a bank of ground-truth \texttt{truck} and \texttt{bus} crops with a minimum side length of 8 pixels, a boundary margin of at least 2 pixels, and an aspect ratio in $[0.3,4.0]$. We augment $40\%$ of frames sampled per sequence (seed 0) by pasting two same-class instances, avoiding same-frame sources whenever possible. Each crop is scaled to the median vehicle height with jitter $[0.85,1.15]$, horizontal flipping ($p=0.5$), and brightness adjustment ($p=0.3$), and is placed in the lower image region with $\operatorname{IoU}<0.10$ within 50 attempts.
For X-Paste, we construct an instance bank with up to 512 instances per class,
using CLIP and segmentation thresholds of 0.28 and 0.45, respectively. We
retain masks with fill ratios between 0.20 and 0.85 and restrict the bounding-box
area ratio to at most 0.55. During placement, we use an IoU threshold of 0.10,
apply scale jittering between 0.85 and 1.15, and perform horizontal flipping and
brightness adjustment with probabilities of 0.5 and 0.3, respectively.
For our method, the 40\% subset is sampled by object count: 50\%, 25\%, and 15\% are assigned to images with zero, one, and multiple objects, respectively, while the remaining 10\% is reserved for hard examples with high object density, small objects, or objects near image boundaries. Some visualizations of generated images for datasets augmentation are shown is Fig.\ref{fig:uavdt-aug}.

\begin{figure}
    \centering
    \includegraphics[width=1.0\linewidth]{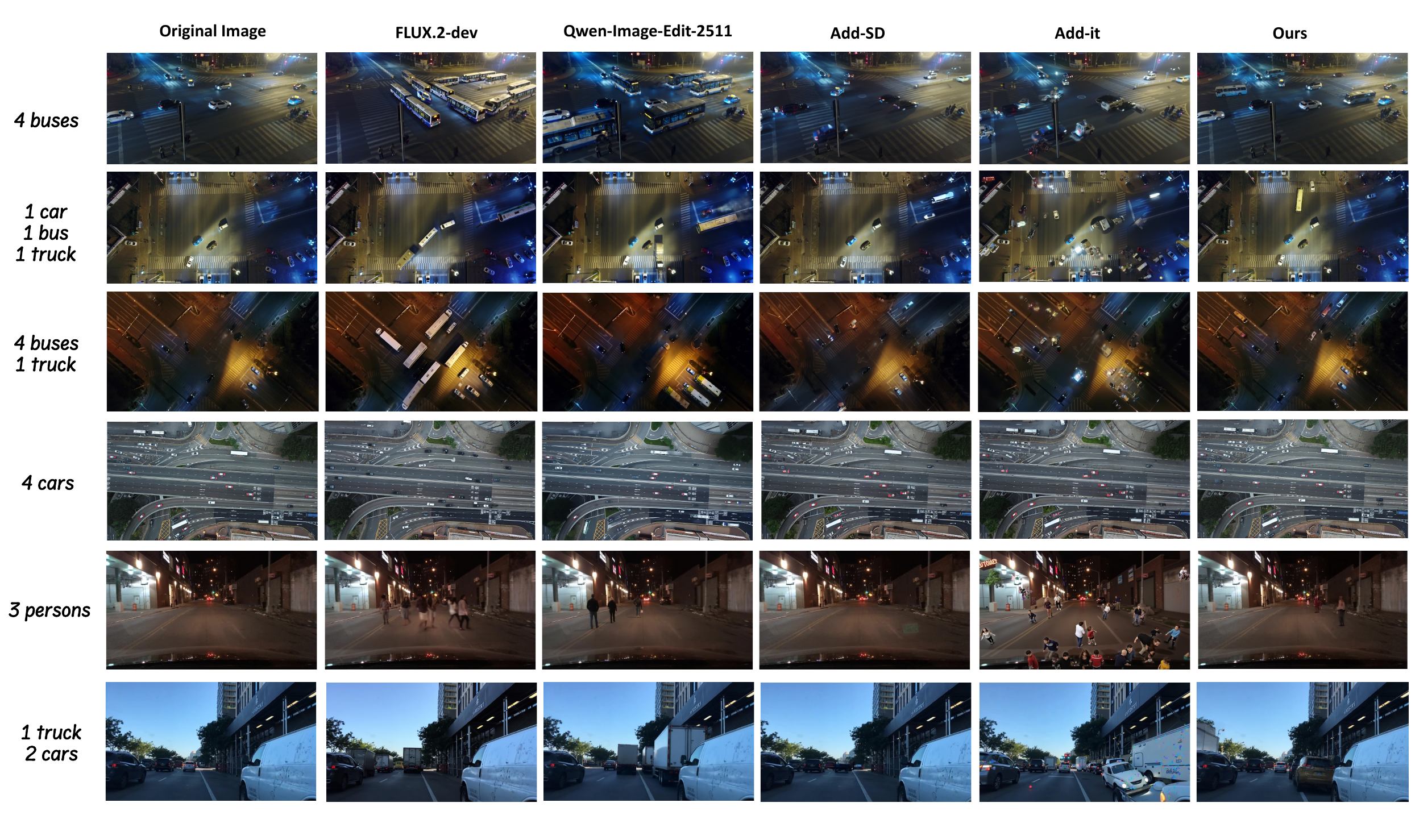}
    \caption{Visualized Comparison between different methods.}
    \label{fig:placeholder}
\end{figure}

\end{document}

%% file: math_commands.tex
\usepackage{amsmath,amsfonts,bm}

\def\eqref#1{equation~\ref{#1}}

\def\1{\bm{1}}

\DeclareMathAlphabet{\mathsfit}{\encodingdefault}{\sfdefault}{m}{sl}
\SetMathAlphabet{\mathsfit}{bold}{\encodingdefault}{\sfdefault}{bx}{n}

